\documentclass[10pt,twocolumn,letterpaper]{article}

\usepackage{cvpr}
\usepackage{times}
\usepackage{epsfig}
\usepackage{graphicx}
\usepackage{amsmath}
\usepackage{amssymb}

\usepackage{caption}
\usepackage{color}
\usepackage{wrapfig}
\usepackage[utf8]{inputenc} % allow utf-8 input
\usepackage[T1]{fontenc}    % use 8-bit T1 fonts
\usepackage{url}            % simple URL typesetting
\usepackage{booktabs}       % professional-quality tables
\usepackage[export]{adjustbox}
\usepackage[pagebackref=true,breaklinks=true,letterpaper=true,colorlinks,bookmarks=false]{hyperref}

\cvprfinalcopy % *** Uncomment this line for the final submission

\def\cvprPaperID{7437} % *** Enter the CVPR Paper ID here

\ifcvprfinal\fi
\begin{document}

% new commands
\newcommand{\ks}[1]{\textcolor{red}{KS: #1}}
\newcommand{\cw}[1]{\textcolor{blue}{CW: #1}}
%\newcommand{\etal}{\textit{et al}. }

%%%%%%%%% TITLE
\title{Affinity Graph Supervision for Visual Recognition}

% \and

\def\mc{\textsuperscript{1}}
\def\ad{\textsuperscript{2}}
\def\iit{\textsuperscript{3}}
\def\adiit{\textsuperscript{2,3}}
% \author{
%   Chu Wang \mc \hspace{0.2cm} Babak Samari \mc \hspace{0.2cm} Vladimir G. Kim \ad \hspace{0.2cm} 
%   Siddhartha Chaudhuri \adiit \hspace{0.2cm} Kaleem Siddiqi \mc \thanks{Corresponding author.} \\
%   \mc School of Computer Science and Center for Intelligent Machines, McGill University\\
%   \ad Adobe Research \\
%   \iit Department of Computer Science and Engineering, IIT Bombay \\
%   \texttt{\small \{chuwang,babak,siddiqi\}@cim.mcgill.ca} ~~~~ \texttt{\small \{vokim,sidch\}@adobe.com} \\
% }
\author{
  Chu Wang \mc ~~~~ Babak Samari \mc ~~~~ Vladimir G. Kim \ad ~~~~ 
  Siddhartha Chaudhuri \adiit ~~~~ Kaleem Siddiqi \mc \thanks{Corresponding author.} \\
  \mc McGill University ~~~~
  \ad Adobe Research ~~~~
  \iit IIT Bombay \\
  \texttt{\small \{chuwang,babak,siddiqi\}@cim.mcgill.ca} ~~~~ \texttt{\small \{vokim,sidch\}@adobe.com} \\
}

\maketitle
\thispagestyle{empty}

%%%%%%%%% ABSTRACT
\begin{abstract}
Affinity graphs are widely used in deep architectures, including graph convolutional neural networks and attention networks. Thus far, the literature has focused on abstracting features from such graphs, while the learning of the affinities themselves has been overlooked. Here we propose a principled method to directly supervise the learning of weights in affinity graphs, to exploit meaningful connections between entities in the data source. Applied to a visual attention network \cite{hu2017relation}, our affinity supervision improves relationship recovery between objects, even without the use of manually annotated relationship labels. We further show that affinity learning between objects boosts scene categorization performance and that the supervision of affinity can also be applied to graphs built from mini-batches, for neural network training. In an image classification task we demonstrate consistent improvement over the baseline, with diverse network architectures and datasets.

\end{abstract}

%%%%%%%%% BODY TEXT
\section{Introduction}
Recent advances in graph representation learning have lead to principled approaches for abstracting features from such structures. In the context of deep learning, graph convolutional neural networks networks (GCNs) have shown great promise \cite{defferrard2016convolutional, kipf2016semi}. %Representing the data source as an affinity graph, where nodes represent entities and edges represent their pairwise affinity, GCNs are capable of abstracting entity-level and structure-level features from the input, leading to the potential of feature learning beyond regular image spaces \cite{monti2016geometric, wang2018local}. 
The affinity graphs in GCNs, whose nodes represent entities in the data source and whose edges represent pairwise affinity, are usually constructed from a predefined metric space and are therefore fixed during the training process \cite{defferrard2016convolutional, kipf2016semi, monti2016geometric, wang2018local}. In related work, self-attention mechanisms \cite{vaswani2017attention} and graph attention networks \cite{velivckovic2017graph} have been proposed. Here, using pairwise weights between entities, a fully connected affinity graph is used for feature aggregation. In contrast to the graphs in GCNs, the parametrized edge weights change during the training of the graph attention module. More recent approaches also consider elaborate edge weight parametrization strategies \cite{jiang2019semi, li2018adaptive} to further improve the flexibility of graph structure learning. However, the learning of edge (attention) weights in the graph is entirely supervised by a main objective loss, to improve performance in a downstream task.
%The general strategy is to shift the edge weights so that the downstream task's performance is improved.

Whereas representation learning from affinity graphs has demonstrated great success in various applications \cite{hu2017relation, zhao2018psanet, wang2018nonlocal, hu2018squeeze, hu2018gather}, little work has been done thus far to directly supervise the learning of affinity weights.
In the present article, we propose to explicitly supervise the learning of the affinity graph weights by introducing a notion of target affinity mass, which is a collection of affinity weights that need to be emphasized. We further propose to optimize a novel loss function to increase the target affinity mass during the training of a neural network, to benefit various visual recognition tasks. The proposed affinity supervision method is generalizable, supporting flexible design of supervision targets according to the need of different tasks. This feature is not seen in the related works, since the learning of such graphs are either constrained by distance metrics \cite{jiang2019semi} or dependent on the main objective loss \cite{vaswani2017attention, velivckovic2017graph, li2018adaptive}.

With the proposed supervision of the learning of affinity weights, a visual attention network \cite{hu2017relation} is able to compete in a relationship proposal task with the present state-of-the-art \cite{relproposal} without any explicit use of relationship labels. Enabling relationship labels provides an additional 25\% boost over \cite{relproposal} in relative terms. This improved relationship recovery is particularly beneficial when applied to a scene categorization task, since scenes are comprised of collections of distinct objects.
% Distinct scenes categories are defined by the collections of distinct object classes that comprise them, and thus emphasizing pairs of co-occurring object classes which affinity supervision allows, is beneficial. 
We also explore the general idea of affinity supervised mini-batch training of a neural network, which is common to a vast number of computer vision and other applications. For image classification tasks we demonstrate a consistent improvement over the baseline, across multiple architectures and datasets. Our proposed affinity supervision method leads to no computational overhead, since we do not introduce additional parameters.
\section{Related Work}

\begin{table}[t]
    \centering
    %\vspace{-1cm}
    %\hspace*{-0.45cm}
    \begin{tabular}{c c}
    %\hline
    \centering
    \includegraphics[width=0.22\textwidth]{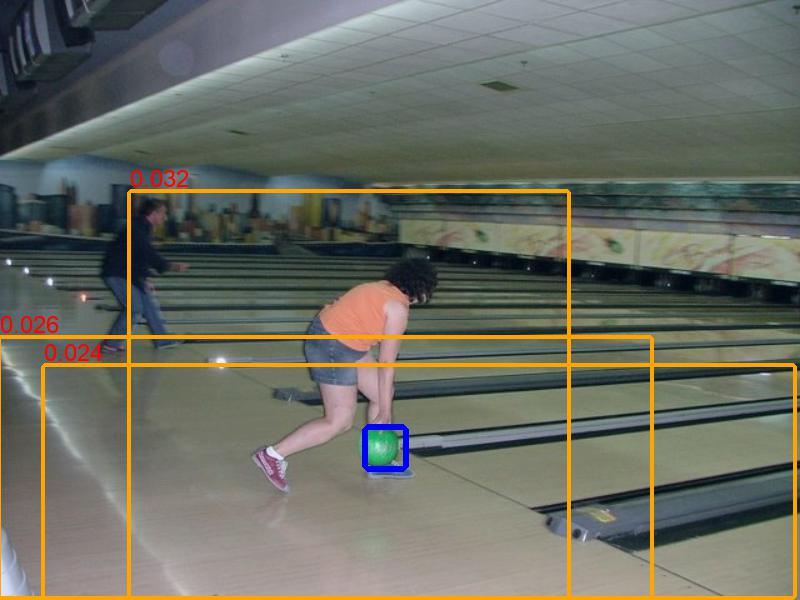} & 
    \includegraphics[width=0.22\textwidth]{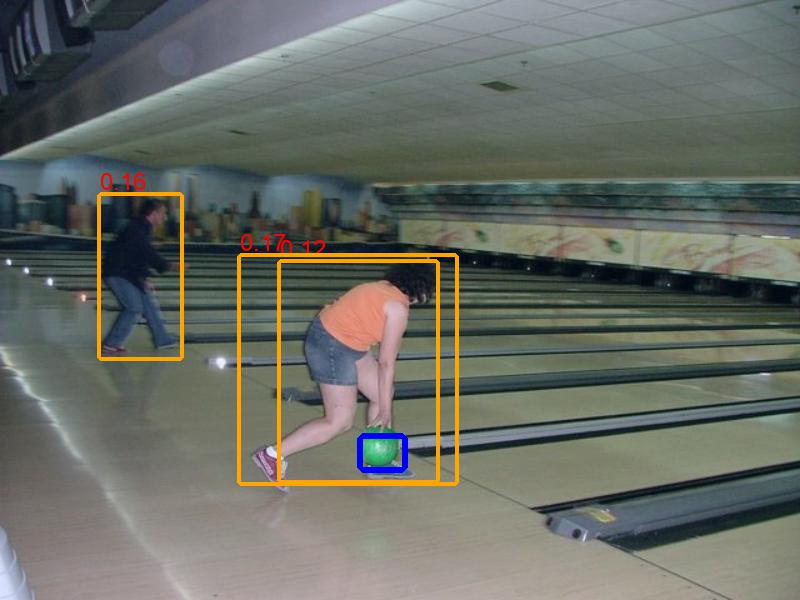} \\
    \includegraphics[width=0.22\textwidth, trim={0 4cm 0cm 0cm}, clip]{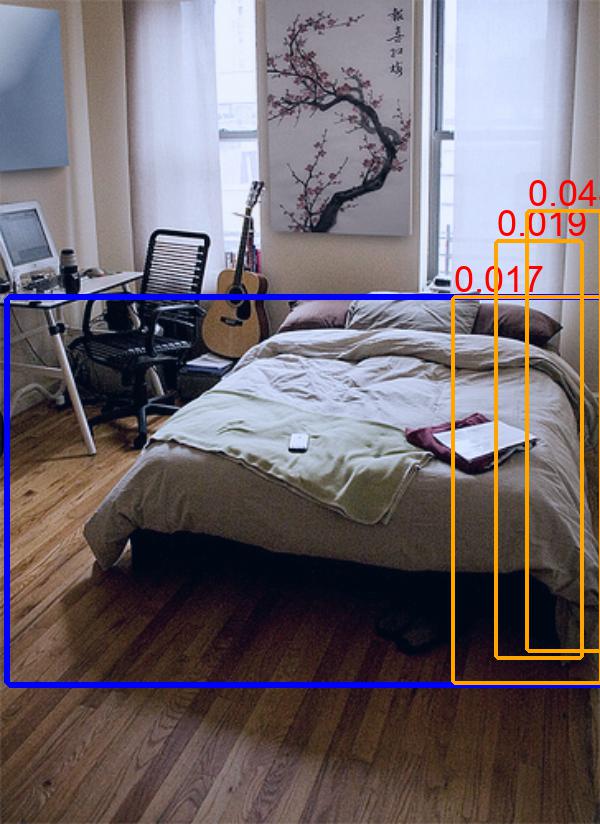} & 
    \includegraphics[width=0.22\textwidth, trim={0 4cm 0cm 0cm}, clip]{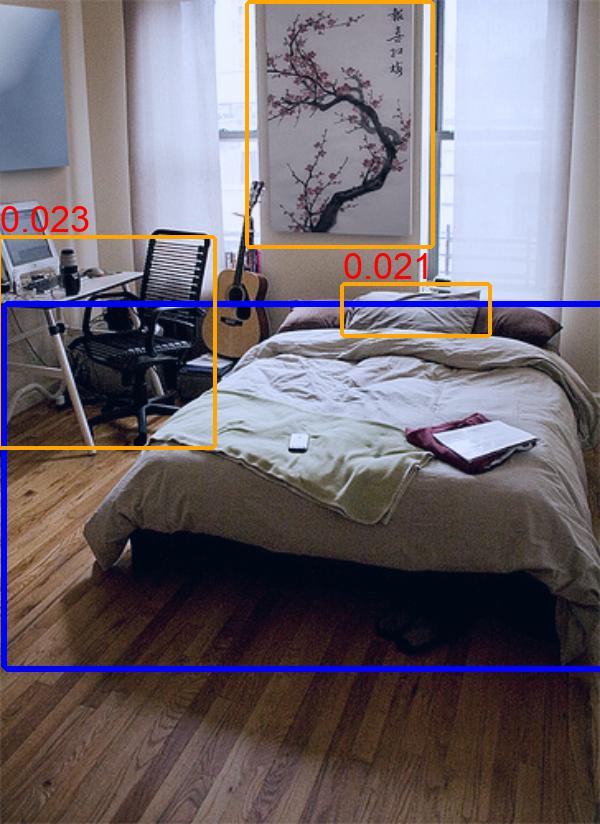} 
    \end{tabular}

    \captionof{figure}{A comparison of recovered relationships on test images, with no relationship annotations used during training. We shows the reference object (blue box), regions with which it learns relationships (orange boxes) and the relationship weights in red text (zoom in on the PDF). Left: baseline visual attention networks \cite{hu2017relation} often recover relationships between a reference object and its immediate surrounding context. Right: our proposed affinity supervision better emphasizes \textit{potential} relationships between distinct and spatially separated objects. %More visualizations of recovered relationships are included in the supplementary material.
    }
    %\vspace{-0.2cm}
    \label{fig:rel_comparison}
\end{table}

\subsection{Graph Convolutional Neural Networks}
In GCNs layer-wise convolutional operations are applied to abstract features in graph structures. Current approaches build the affinity graph from a predefined input \cite{defferrard2016convolutional, monti2016geometric, wang2018local} or embedding space \cite{hamilton2017inductive, vaswani2017attention}, following which features are learned using graph based filtering in either the spatial or spectral domain.
Little work has been carried out so far to directly learn the structure of the affinity graph itself. In this article, we propose a generic method for supervising the learning of pairwise affinities in such a graph, without the need for additional ground truth annotations. 

\subsection{Visual Attention Networks} 
Attention mechanisms, first proposed in \cite{vaswani2017attention}, have been successfully applied to a diverge range of computer vision tasks \cite{hu2017relation, zhao2018psanet, wang2018nonlocal}. In the context of object detection \cite{hu2017relation}, the attention module uses learned pairwise attention weights between region proposals, followed by per region feature aggregation, to boost object detection. The learned attention weights do not necessarily reflect relations between entities in a typical scene. In fact, for a given reference object (region), relation networks \cite{hu2017relation} tend to predict high attention weights with scaled or shifted bounding boxes surrounding the same object instance (Figure \ref{fig:rel_comparison}). 

A present limitation of visual attention networks %\cite{hu2017relation, zhao2018psanet, wang2018nonlocal} %in various applications 
is their minimization of only the main objective loss during training \cite{hu2017relation, zhao2018psanet, wang2018nonlocal}, without any direct supervision of attention between entities. Whereas attention based feature aggregation has been shown to boost performance for general vision tasks \cite{hu2018squeeze, hu2018gather}, the examples in Figure \ref{fig:rel_comparison} provide evidence that relationships between distinct entities may not be sufficiently captured. In this paper we address this limitation by directly supervising the learning of attention. An affinity graph is first build from the pair-wise attention weights and a novel target affinity mass loss is then applied to guide the learning of attention between distinct objects, allowing the recovery of more plausible relationships. 

\subsection{Mini-batch Training}
The training of a neural network often requires working with mini-batches of data, because typical datasets are too large for present architectures to handle. The optimization of mini-batch training is thus a research topic in its own right. Much work has focused on improving the learning strategies, going beyond stochastic gradient decent (SGD), including \cite{qian1999momentum, duchi2011adaptive, rmsprop, kingma2014adam}. In addition, batch normalization \cite{ioffe2015batch} has shown to improve the speed, performance, and stability of mini-batch training, via the  normalization of each neuron's output to form a unified Gaussian distribution across the mini-batch. 

In the present article we show that our affinity supervision on a graph built from mini-batch features can benefit the training of a neural network. By increasing the affinity (similarity) between mini-batch entries that belong to the same category, performance in image classification on a diverse set of benchmarks, is consistently improved. We shall discuss mini-batch affinity learning in more detail in Section \ref{sec:batch_affinity}.

\section{Affinity Graph Supervision}\label{sec:aff_sup}
We now introduce our approach to supervising the weights in an affinity graph. Later we shall cover two applications: affinity supervision on visual attention networks (built on top of Relation Networks \cite{hu2017relation}) in Section \ref{sec:att_nets} and affinity supervision on a batch similarity graph in Section \ref{sec:batch_affinity}. 

\subsection{Affinity Graph}\label{sec:affinity_graph}
We assume that there are $N$ entities generated by a feature embedding framework, for example, a region proposal network (RPN) together with ROI pooling on a single image \cite{fasterRCNN}, or a regular CNN applied over a batch of images. 
Let $\mathbf{f}^i$ be the embedding feature for the $i$-th entity. We define an affinity function $\mathcal{A}$ which computes an affinity weight between a pair of entities $m$ and entity $n$, as
\begin{equation}
\omega^{mn} = \mathcal{A}(\mathbf{f}^m, \mathbf{f}^n).
\label{eq.att_fn}
\end{equation}
A specific form of this affinity function applied in attention networks \cite{hu2017relation, vaswani2017attention} is reviewed in Section \ref{sec:att_nets}, and another simple form of this affinity function applied in batch training is defined in section \ref{sec:batch_affinity}. 

We now build an affinity graph $G$ whose vertices $m$ represent entities in the data source with features $\mathbf{F}_{in} = \{\mathbf{f}^m\}$ and whose edge weights $\{\omega^{mn}\}$ represent pairwise affinities between the vertices. We define the graph adjacency matrix for this affinity graph as the $N \times N$ matrix $\mathcal{W}$ with entries $\{\omega^{mn}\}$.
We propose to supervise the learning of $\mathcal{W}$ so that those matrix entries $\omega^{mn}$ selected by a customized supervision target matrix $\mathcal{T}$ will increase, thus gaining emphasis over the other entries.

\subsection{Affinity Target $\mathcal{T}$} \label{sec:suptarget}
We now explain the role of a supervision target matrix $\mathcal{T}$ for affinity graph learning. In general, $\mathcal{T} \in \mathbf{R}^{N \times N}$ with
\begin{equation}
  \mathcal{T}[i,j] =
  \begin{cases}
    1 \quad \textrm{if } (i,j) \in \mathcal{S} \\
    0 \quad \textrm{otherwise},
  \end{cases}
\end{equation}
where $\mathcal{S}$ stands for a set of possible connections between entities in the data source. 

\begin{figure*}[t]
    % \left
    \centering
    \vspace{-1.5cm}
    \includegraphics[width=1.0\textwidth, trim={0cm 0cm 0cm 0cm}, clip]{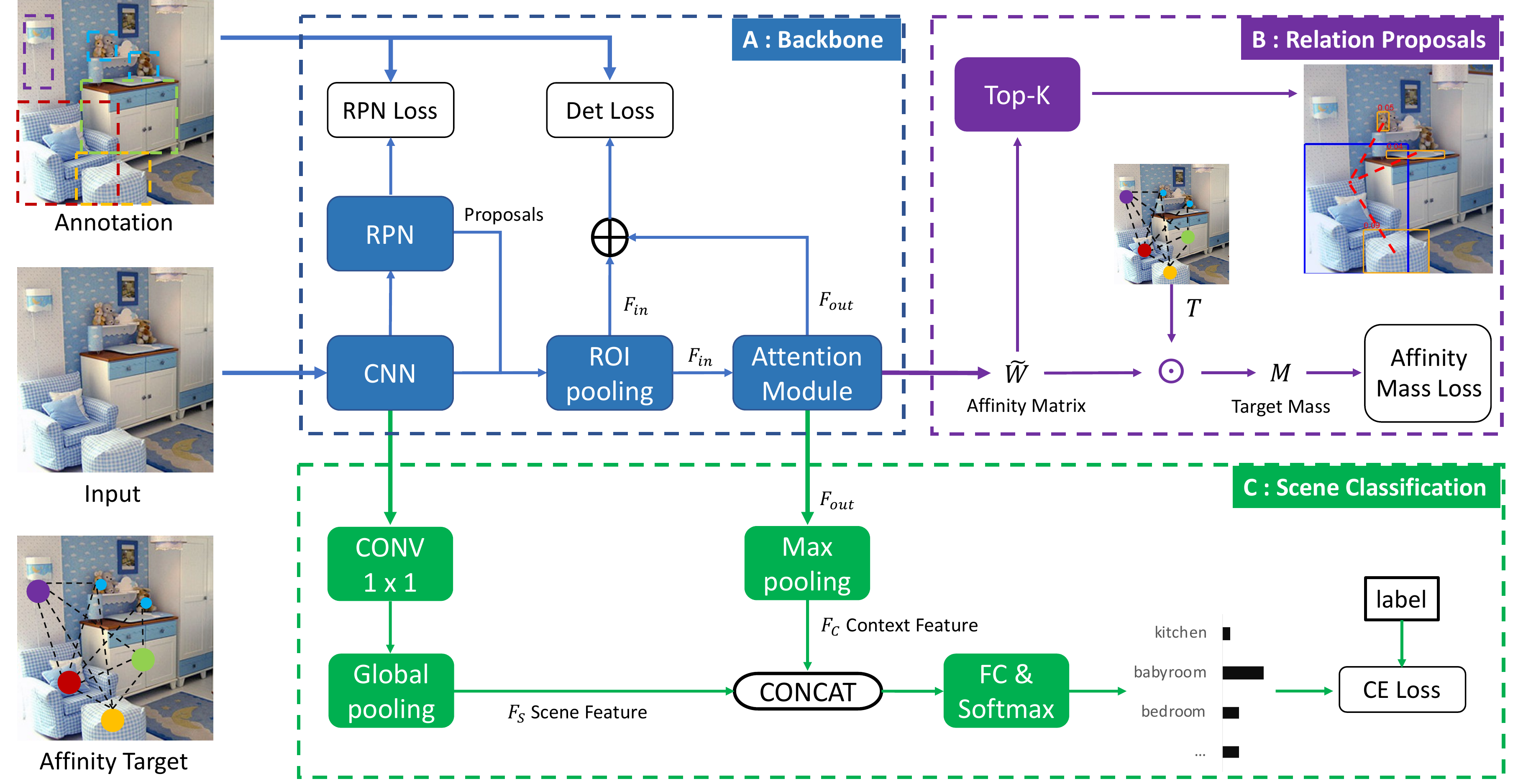}
    \caption{An overview of our affinity graph supervision in visual attention networks, in application to two tasks. The blue dashed box surrounds the visual attention network backbone, implemented according to Relation Networks \cite{hu2017relation}. The purple dashed box highlights our core component for affinity learning and for relation proposal generation. The green dashed box surrounds the branch for scene categorization. An example affinity target is visualized in the bottom left corner, with solid circles representing ground truth objects colored by their class. The dashed lines between pairs of solid circles give rise to a value of 1 for the corresponding entry in matrix $\mathcal{T}$. See the text in Section \ref{sec:attn_target} for a detailed description. A detailed illustration of the attention module is in the supplementary material. }
    \label{fig:net_architecture}
\end{figure*}

\paragraph{Target Affinity Mass} 
We would like $\mathcal{W}$ to have higher weights at those entries where $\mathcal{T}[i,j] = 1$, to place emphasis on the entries that are selected by the supervision target.
We capture this via a notion of {\em target affinity mass} $\mathcal{M}$ of the affinity graph, defined as 
\begin{equation}
\mathcal{M} = \sum  \mathcal{\tilde W} \odot  \mathcal{T},
\label{eq:target_mass}
\end{equation}
where $\mathcal{\tilde W} = softmax( \mathcal{W} ) $ is a matrix-wise softmax operation. A study on affinity mass design is available in the supplementary material.

\subsection{Affinity Mass Loss $\mathcal{L}_{G}$} 
\label{sec:massloss}
We propose to optimize the learning of the parameters $\theta$ of a neural network to achieve 
\begin{equation}\label{eq:mass_opt_goal}
    \max_{\theta} \quad \mathcal{M}.
\end{equation}
Our aim is to devise a strategy to maximize $\mathcal{M}$ with an empirically determined choice of loss form. There are several loss forms that could be considered, including smooth $L1$ loss, $L2$ loss, and a focal loss variant. Defining $x =  1 - \mathcal{M} \in [0, 1]$, we define losses 
\begin{equation}
    L_2(x) = x^2
\end{equation}
and
\begin{equation}
  \textrm{smooth}_{L_1}(x) =
  \begin{cases}
    x^2 \quad \textrm{if } \quad |x| < 0.5\\
    |x| - 0.25 \quad \textrm{otherwise}.
  \end{cases}
\end{equation}
The focal loss on $\mathcal{M}$ is a negative log likelihood loss, weighted by the focal normalization term proposed in \cite{focalloss}, which is defined as 
\begin{equation}
\mathcal{L}_{G} = L_{\text{focal}}(\mathcal{M})  = - (1 - \mathcal{M})^\gamma \log (\mathcal{M}).
\label{eq.aff_loss}
\end{equation} 
The focal term $(1- \mathcal{M})^\gamma$ \cite{focalloss} helps narrow the gap between well converged affinity masses and those that are far from convergence. 

Empirically, we have found that the focal loss variant gives the best results in practice, as described in the ablation study reported in Section \ref{sec:exp_ablation}. The choice of the $\gamma$ term depends on the particular tasks, so we provide experiments to justify our choices in Section \ref{sec:exp_ablation}. 

\subsection{Optimization and Convergence of $\mathcal{L}_{G}$} \label{sec:mass_optimization}
The minimization of the affinity mass loss $\mathcal{L}_{G}$ places greater emphasis on entries 
in $\mathcal{W}$ which correspond to ground truth connections in $\mathcal{S}$, through network training. However, when optimized in conjunction with a main objective loss, which could be an object detection loss $\mathcal{L}_{main} = \mathcal{L}_{det} + \mathcal{L}_{rpn}$ in visual attention networks or a cross entropy loss $\mathcal{L}_{main} = \mathcal{L}_{class}$ in mini-batch training, a balance between $\mathcal{L}_{main}$ and $\mathcal{L}_{G}$ is required. 
The total loss can be written as
\begin{equation}
    \mathcal{L} = \mathcal{L}_{main} + \lambda \mathcal{L}_{G}.
\end{equation}
Empirically, we choose $\lambda = 0.01$ for visual attention networks and for mini-batch training, we choose $\lambda = 0.1$. Figure \ref{fig:batch_aff_analysis} demonstrates the convergence of the target mass, justifying the effectiveness of using loss $\mathcal{L}_{G}$ in the optimization of equation \ref{eq:mass_opt_goal}.
\section{Affinity in Attention Networks} \label{sec:att_nets}
We review the computation of attention weights in %the dot product attention module of 
\cite{vaswani2017attention}, given a pair of nodes from the attention graph defined in Section \ref{sec:affinity_graph}. Let an entity node $m$ consist of its feature embedding, defined as $\mathbf{f}^m$. The collection of input features of all the nodes then becomes $\mathbf{F}_{in} =\{ \mathbf{f}^m \}$.
Consider node $m$ as a reference object with the attention weight $\tilde{\omega}^{mn}$ indicating its affinity to a surrounding entity node $n$. This affinity is computed as a softmax activation over the scaled dot products $\omega^{mn}$ defined as: 
\begin{equation}
\tilde{\omega}^{mn} = \frac{\exp(\omega^{mn})}{\sum_{k} \exp(\omega^{kn})}, \quad
\omega^{mn} = \frac{<W_K{\mathbf{f}}^{m}, W_Q \mathbf{f}^{n}>}{\sqrt{d_k}}.
\label{eqn:object_relation_weight}
\end{equation}
Both $W_K$ and $W_Q$ are matrices and so this linear transformation projects the embedding features $\mathbf{f}^m$ and $\mathbf{f}^n$ into metric spaces to measure how well they match. The feature dimension after projection is $d_k$. With the above formulation, the attention graph affinity matrix is defined as $\mathcal{W} = \{{\omega}^{mn} \}$. For a given reference entity node $m$, the attention module also outputs a weighted aggregation of $m$'s neighbouring nodes' features, which is 
\begin{equation} \label{eq:att_aggregate}
\mathbf{f}_{out}^m = \sum_n \tilde{\omega}^{mn} f^n.
\end{equation}
The set of eature outputs for all nodes is thus defined as $\mathbf{F}_{out} = \{ \mathbf{f}_{out}^m \}$. Additional details are provided in \cite{vaswani2017attention, hu2017relation}.

\subsection{Affinity Target Design} \label{sec:attn_target}
For visual attention networks, we want our attention weights to focus on relationships between objects from different categories, so for each entry $\mathcal{T}[a,b]$ of the supervision target matrix $\mathcal{T}$, we assign $\mathcal{T}[a,b] = 1$ only when: 
\begin{enumerate}
    \item proposal $a$ overlaps with ground truth object $\alpha$'s bounding box with intersection over union $> 0.5$.
    \item  proposal $b$ overlaps with ground truth object $\beta$'s bounding box with intersection over union $> 0.5$.
    \item ground truth objects $\alpha$ and $\beta$ are two different objects coming from different classes.
\end{enumerate}
Note that NO relation annotation is required to construct such supervision target.

We choose to emphasize relationships between exemplars from different categories in the target matrix, because this can provide additional contextual features in the attention aggregation (equation \ref{eq:att_aggregate}) for certain tasks. Emphasizing relationships between objects within the same category might be better suited to modeling co-occurence. We provide a visualization of the affinity target and additional studies, in the supplementary material. We now discuss applications that could benefit from affinity supervision of the attention weights: object detection, relationship proposal generation, and scene categorization.

\subsection{Object Detection and Relationship Proposals} \label{sec:detection}
In Figure \ref{fig:net_architecture} (part A to part B) we demonstrate the use of attention networks for object detection and relationship proposal generation. Here part A is identical to Relation Networks \cite{hu2017relation}. The network is end-to-end trainable with detection loss, RPN loss and the target affinity mass loss. In addition to the ROI pooling features $\mathbf{F}_{in} \in \mathcal{R}^{N_{obj} \times 1024}$ from the Faster R-CNN backbone of \cite{fasterRCNN}, contextual features $\mathbf{F}_{out}$ from attention aggregation are applied to boost detection performance. 
The final feature descriptor for the detection head is $\mathbf{F} + \mathbf{F}_c$, following \cite{hu2017relation}. In parallel, the attention matrix output $\mathcal{W} \in \mathcal{R}^{N \times N}$ is used to generate relationship proposals by finding the top K weighted pairs in the matrix. 

\subsection{Scene Categorization} \label{sec:scene_net}
In Figure \ref{fig:net_architecture} (part A to part C) we demonstrate an application of visual attention networks to scene categorization. Since there are no bounding box annotations in most scene recognition datasets, we adopt a visual attention network (described in the previous section), pretrained on the MSCOCO dataset, in conjunction with a new scene recognition branch (part C in Figure \ref{fig:net_architecture}), to perform scene recognition. From the CNN backbone, we apply an additional $1 \times 1$ convolution layer, followed by a global average pooling to acquire the scene level feature descriptor $\mathbf{F}_s$. The attention module takes as input the object proposals' visual features $\mathbf{F}_{in}$, and outputs the aggregation result as the scene contextual feature $\mathbf{F}_c$. The input to the scene classification head thus becomes $\mathbf{F}_{meta} = concat(\mathbf{F}_s, \mathbf{F}_c)$, and the class scores are output. In order to maintain the learned relationship weights from the pre-trained visual attention network, which helps encode object relation context in the aggregation result $\mathbf{F}_{out}$, we fix the parameters in part A (blue box), but make all other layers in part C trainable.

\section{Affinity in mini-Batch}
\label{sec:batch_affinity}
Moving beyond the specific problems of object detection, relationship proposal generation and scene categorization, we now turn to a more general application of affinity supervision, that of mini-batch training in neural networks. Owing to the large size of most databases and limitations in memory, virtually all deep learning models are trained using mini-batches. We shall demonstrate that emphasizing pairwise affinities between entities during training can boost performance for a variety of image classification tasks.

\subsection{Affinity Graph} We consider image classification over a batch of $N$ images, processed by a convolutional neural network (CNN) to generate feature representations. Using the notation in Section \ref{sec:aff_sup}, we denote the feature vectors of this batch of images as $\mathbf{F}_{in} = \{\mathbf{f}^i \}$, where $i \in {1...N}$ is the image index in the batch. We then build a batch affinity graph $G$ whose nodes represent images, and the edge $\omega^{mn}\in \mathcal{W}$ encode pairwise feature similarity between node $m$ and $n$. 

\paragraph{Distance Metric.} A straightforward $L2$ distance based measure \footnote{More elaborate distance metrics could also be considered, but that is beyond the focus of this article.} can be applied to compute the edge weights as
\begin{equation}
    \omega^{mn} = \mathcal{A}(f^m, f^n) = -\frac{\| f^m - f^n \|_{2}^{2}}{2 }.
\end{equation}

\begin{figure}[t]
    % \left
    \centering
    \includegraphics[width=0.5\textwidth, trim={0cm 0cm 0cm 0cm}, clip]{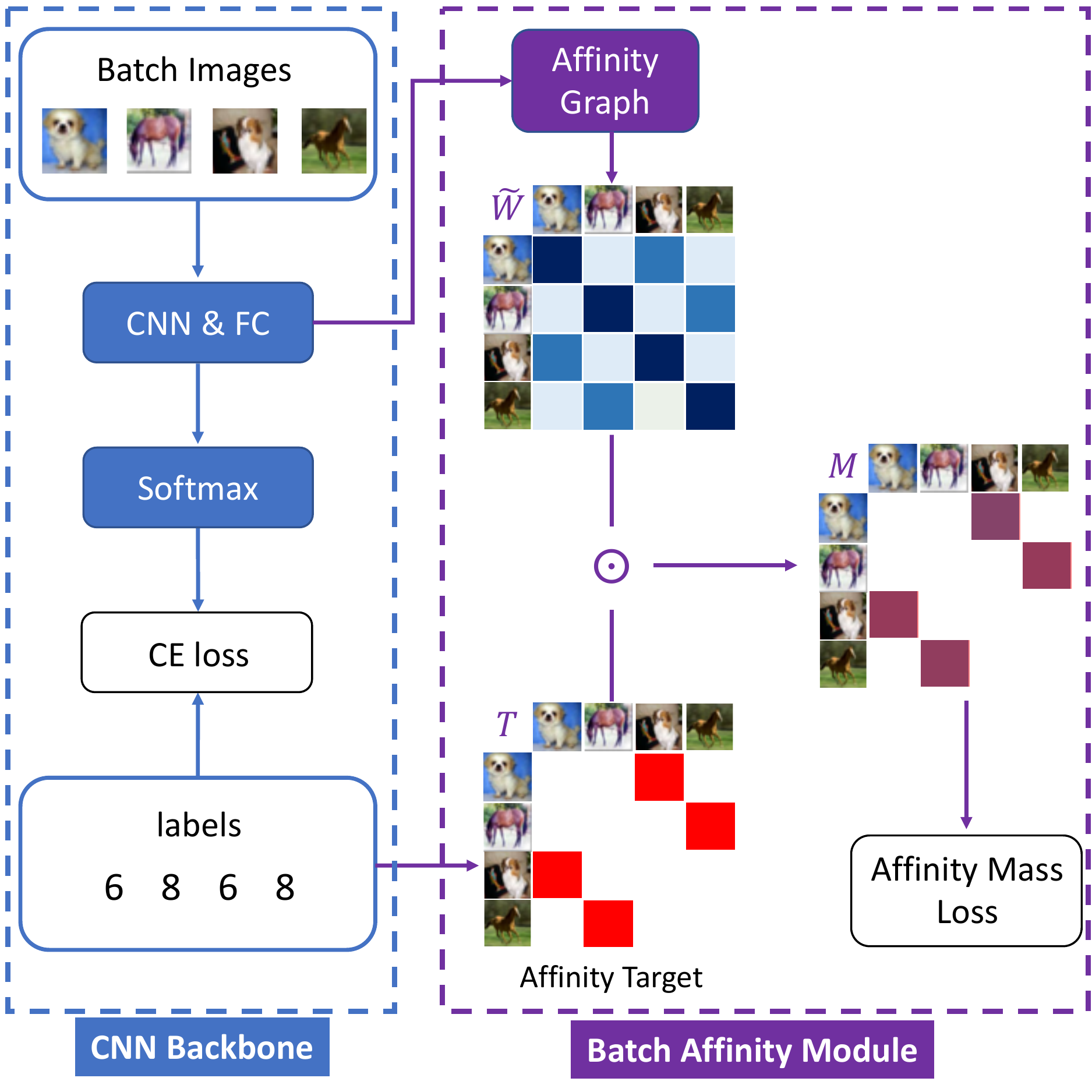}
    \caption{An overview of our affinity graph supervision in mini-batch training of a standard convolutional neural network. Blue box: CNN backbone for image classification. Purple box: Affinity supervision module for mini-batch training. The colored tiles represent entries of the affinity matrix $\tilde{\mathcal{W}}$ and target $\mathcal{T}$, where a darker color denotes a larger numerical value. Minimization of the affinity mass loss aims to increase the value of the purple squares representin entries in mass $\mathcal{M}$ (see equation \ref{eq:target_mass}). 
    % See text in section \ref{sec:batch_structure} for details regarding each step. 
    }
    \label{fig:batch_affinity}
\end{figure}

\subsection{Affinity Target Design} \label{sec:batch_target}
In the mini-batch training setting, we would like feature representations from the same class to be closer to each other in a metric space, with those from different classes being spread apart.
To this end, we build the affinity target matrix $\mathcal{T}$ as follows. For each entry $\mathcal{T}[a,b]$ in the matrix, we assign $\mathcal{T}[a,b] = 1$ only when mini-batch node $a$ and $b$ belong to the same category. Thus, the affinity target here selects those entries in $\mathcal{W}$ which represent pairwise similarity between images from the same class. During the optimization of the affinity mass loss (defined in Section \ref{sec:massloss}), the network will increase the affinity value from the entries in $\mathcal{W}$ selected by $T$, while suppressing the other ones. This should in principle leads to improved representation learning and thus benefit the underlying classification task. 

\subsection{Overview of Approach} \label{sec:batch_structure}
A schematic overview of our mini-batch affinity learning approach is presented in Figure \ref{fig:batch_affinity}. Given a batch of $N$ images, we first generate the feature representations $\mathbf{F}_{in}$ from a CNN followed by fully connected layers. We then send $\mathbf{F}_{in}$ to an affinity graph module, which contains a pair-wise distance metric computation followed by a matrix-wise softmax activation, to acquire the affinity graph matrix $\Tilde{\mathcal{W}}$. Next, we built the affinity target matrix $\mathcal{T}$ from the image category labels following Section \ref{sec:batch_target}. An element-wise multiplication with $\Tilde{\mathcal{W}}$ is used to acquire the target affinity mass $\mathcal{M}$, which is used in computing the affinity mass loss. During training, the network is optimized by both cross entropy loss $\mathcal{L}_{class}$ and the target affinity loss $\mathcal{L}_{G}$, using the balancing scheme discussed in Section \ref{sec:mass_optimization}.

\section{Experiments} \label{sec:exp}
\subsection{Datasets} \label{sec:data_train}
%We evaluate the affinity graph supervision using the following datasets. \\
\noindent \textbf{VOC07:}, which is part of the PASCAL VOC detection dataset \cite{voc}, with 5k images in trainval and 5k in test set. We used this trainval/test split for model ablation purposes. \\ 
\textbf{MSCOCO:} which consists of 80 object categories \cite{mscoco}. We used the 30k validation images for training and the 5k ``minival'' images for testing, which is common practice \cite{hu2017relation}. \\
\textbf{Visual Genome:} which is a large relationship understanding benchmark \cite{visualgenome}, consisting of 150 object categories and human annotated relationship labels between objects. We used 70k images for training and 30K for testing, as in the scene graph literature \cite{neuralmotifs,xu2017scenegraph}.\\ 
\textbf{MIT67:} which is a scene categorization benchmark with 67 scene categories, with 80 training images and 20 test images  in each category \cite{mit67}. We used this official split. \\
\textbf{CIFAR10/100:} which is a popular benchmark dataset containing 32 by 32 tiny images from 10 or 100 categories \cite{cifar}. We used the official train/test split and we randomly sampled 10\% of train set to form a validation set. 
\\
\textbf{Tiny Imagenet:} which is a simplified version of the ILSVRC 2012 image classification challenge \cite{deng2009imagenet} containing 200 classes \cite{tinyimagenet} with 500 training images and 50 validation images in each class. We used the official validation set as the test set since the official test set is not publicly available. For validation, we randomly sample 10\% of the training set.

\subsection{Network Training Details} \label{sec:net_training}
\paragraph{Visual Attention Networks.} We first train visual attention networks \cite{hu2017relation} end-to-end, using detection loss, RPN loss and affinity mass loss (Figure \ref{fig:net_architecture} parts A and B). The loss scale for affinity loss is chosen to be $0.01$ as discussed in Section \ref{sec:mass_optimization}. Upon convergence, the network can be directly applied for object detection and relationship proposal tasks. For scene categorization, we first acquire a visual attention network that is pretrained on the COCO dataset, and then use the structural modification in Section \ref{sec:scenecategorization} (Figure \ref{fig:net_architecture} parts A and C) to fine tune it on the MIT67 dataset. Unless stated otherwise, all visual attention networks are based on a ResNet101 \cite{he2016deep} architecture, trained with a batch size of 2 (images), using a learning rate of $5e-4$ which is decreased to $5e-5$ after 5 epochs. There are 8 epochs in total for the each training session. We apply stochastic gradient decent (SGD) with momentum optimizer and set the momentum to $0.9$. We evaluate the model at the end of 8 epochs on the test set to report our results. 

\paragraph{Mini-batch Affinity Supervision.} We applied various architectures including ResNet-20/56/110 for CIFAR and ResNet-18/50/101 for tiny ImageNet, as described in \cite{he2016deep}. \footnote{The network architectures are exactly the same as those in the original ResNet paper \cite{he2016deep}.} 
The CIFAR networks are trained for 200 epochs with a batch size of 128. We set the initial learning rate to 0.1 and reduce it by a factor of 10 at epochs 100 and 150, respectively. The tiny ImageNet networks are trained for 90 epochs with a batch size of 128, an initial learning rate of 0.1, and a factor of 10 reduction at epochs 30 and 60. For all experiments in mini-batch affinity supervision, the SGD optimizer with momentum is applied, with the weight decay and momentum set to $5e-4$ and $0.9$.
For data augmentation during training, we have applied random horizontal flipping. \footnote{For the CIFAR datasets, we also applied 4-pixel padding, followed by $32 \times 32$ random cropping after horizontal flipping, following \cite{he2016deep}.} 
During training we save the best performing model on the validation set, and report test set performance on this model.

\begin{table*}[!ht]
	\centering
	% \hspace{-0.5cm}
	\scalebox{1.0}{
	\begin{tabular}{@{}l||c | c||c | c || c|c|c}
		\hline
		VOC07 Ablation    & F-RCNN \cite{fasterRCNN} & RelNet \cite{hu2017relation} & smooth L1        & L2             & $\gamma=$ 0   & $\gamma=$ 2 & $\gamma=$ 5         \\ \hline
		mAP@all (\%)   &  47.0 & 47.7 $\pm$ 0.1 & 48.0 $\pm$ 0.1 & 47.7 $\pm$ 0.2 & 47.9 $\pm$ 0.2 & 48.2 $\pm$ 0.1          & \textbf{48.6 $\pm$ 0.1} \\
		mAP@0.5 (\%)   &  78.2 & 79.3 $\pm$ 0.2 & 79.6 $\pm$ 0.2 & 79.7 $\pm$ 0.2 & 79.4 $\pm$ 0.1 & 79.9 $\pm$ 0.2          & \textbf{80.0 $\pm$ 0.2} \\
		recall@5k (\%) &  - & 43.5  & 60.3 $\pm$ 0.3 & 64.6 $\pm$ 0.5 & 62.1 $\pm$ 0.3 & \textbf{69.9 $\pm$ 0.3} & 66.8 $\pm$ 0.2 \\
		\hline
	\end{tabular}
	}
	\caption{An ablation study on loss functions comparing against the baseline faster RCNN \cite{fasterRCNN} and Relation Networks \cite{hu2017relation}, using the VOC07 database. The results are reported as percentages (\%) averaged over 3 runs. The relationship recall metric is also reported with ground truth relation labels constructed as described in Section \ref{sec:attn_target}, {\em using only object class labels}. 
	}
	\label{tab:loss_ablation}
\end{table*}

\begin{table*}[!ht]
	\centering
	\scalebox{1.0}{
	\begin{tabular}{@{}l|c|c|c|c|c}
		\hline
		MIT67 & CNN & CNN & CNN + ROIs & CNN + Attn & CNN + Attn + $\mathcal{L}_{G}$ \\ \hline
		Pretraining  & Imgnet  & Imgnet+COCO  & Imgnet+COCO & Imgnet+COCO & Imgnet+COCO   \\ \hline
		Features & $\mathbf{F}_S$  & $\mathbf{F}_S$  & $\mathbf{F}_S, max(\mathbf{F_{in}})$ & $\mathbf{F}_S, \mathbf{F}_C$  & $\mathbf{F}_S, \mathbf{F}_C$\\ \hline
		Accuracy (\%) & 75.1  & 76.8  & 78.0 $\pm$ 0.3 & 77.1 $\pm$ 0.2 & \textbf{80.2 $\pm$ 0.3}   \\
		\hline
	\end{tabular}
	}
	
	\caption{MIT67 Scene Categorization Results, averaged over 3 runs. A visual attention network with affinity supervision gives the best result (the boldfaced entry), with an improvement over a non-affinity supervised version (4-th column) and the baseline methods (columns 1 to 3). See the text in Section \ref{sec:scenecategorization} for details. $F_s$, $F_c$ and $F_{in}$ are described in Section \ref{sec:scene_net}. 
	}
	\label{tab:scenecategorization}
\end{table*}

\subsection{Tasks and Metrics}\label{sec:evaluation}
We evaluate affinity graph supervision on the following tasks, using the associated performance metrics.\\
{\bf Relationship Proposal Generation.} We evaluate the learned relationships on the Visual Genome dataset, using a recall metric which measures the percentage of ground truth relations that are covered in the predicted top K relationship list, which is consistent with \cite{relproposal,neuralmotifs,xu2017scenegraph}. \\
{\bf Classification.} For the MIT67, CIFAR10/100 and Tiny ImageNet evaluation, we use classification accuracy.\\
{\bf Object Detection.} For completeness we also evaluate object detection on VOC07, using mAP (mean average precision) as the evaluation metric \cite{voc, mscoco}. Additional detection results on MSCOCO are in the supplementary material.

\subsection{Ablation Study on Loss Functions} \label{sec:exp_ablation}
We first carry out ablation studies to examine different loss functions for optimizing the target affinity mass $\mathcal{M}$ as well as varying focal terms $r$, as introduced in Section \ref{sec:massloss}. The results in Table \ref{tab:loss_ablation} show that focal loss is in general better than smooth L1 and L2 losses, when supervising the target mass. In our experiments on visual attention networks, we therefore apply focal loss with $\gamma=2$, which empirically gives the best performance in terms of recovering relationships while still maintaining a good performance in detection task. The results in Table \ref{tab:loss_ablation} serve solely to determine the best loss configuration. Here  we do not claim improvement on detection tasks. The results of additional tests using ablated models will be updated in the supplementary material.

\subsection{Relationship Proposal Task}
\label{sec:relationshipproposal} 
\begin{figure}[ht]
    %\hspace{-0.5cm}
    \centering
    \includegraphics[width=.45\textwidth, trim={0cm 0cm 1.5cm 0cm}, clip]{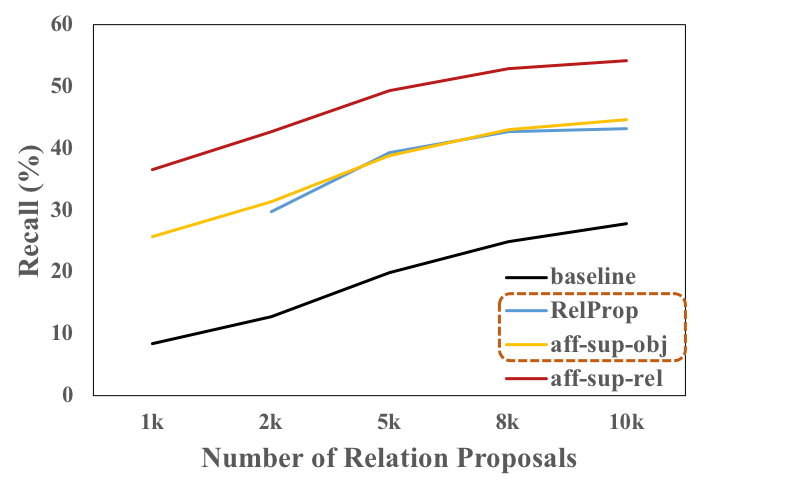}
    \caption{We show the percentage of the true relations that are in the top $K$ retrieved relations, with varying $K$, in a relation proposal task. We compare a baseline network (black), Relation Proposal Networks \cite{relproposal} (blue), our affinity supervision using object class labels (but no explicit relations) (orange) and our affinity supervision with ground truth relation labels (red). We match the state of the art {\em with no ground truth relation labels used} (the overlapping blue and orange curves). We out perform the state of the art by a large margin (25\% in relative terms) when ground truth relations are used.
    }
    %\vspace{-0.5cm}
    \label{fig:rel_eval}
\end{figure}

Figure \ref{fig:rel_eval} compares the relationships recovered on the Visual Genome dataset, by a default visual attention network ``baseline'' model (similar to \cite{hu2017relation}), our affinity supervised network with affinity targets built using only object class labels ``aff-sup-obj'' (see Section \ref{sec:attn_target}), and an affinity target built from human annotated ground truth relation labels ``aff-sup-rel''. We also include the reported recall metric from Relationship Proposal Networks \cite{relproposal}, which is a state-of-the-art level one-stage relationship learning network with strong supervision, using ground truth relationship annotations. Notably, our proposed affinity mass loss does not require potentially costly human annotated relationship labels for learning (only object class labels were used) and yet it achieves the same level of performance as the present state-of-the-art \cite{relproposal} (the blue curve in Figure \ref{fig:rel_eval}). When supervised with a target built from the ground truth relation labels instead of the object labels, we considerably outperform relation proposal networks (by 25\% in relative terms for all K thresholds) with this recall metric (the red curve).

\begin{table*}[!ht]
    \centering
    %\vspace{-0.5cm}
    \begin{tabular}{c c c c}
    \centering
  \includegraphics[width=0.23\textwidth, trim={2.5cm 1cm 2.5cm 1cm}, clip]{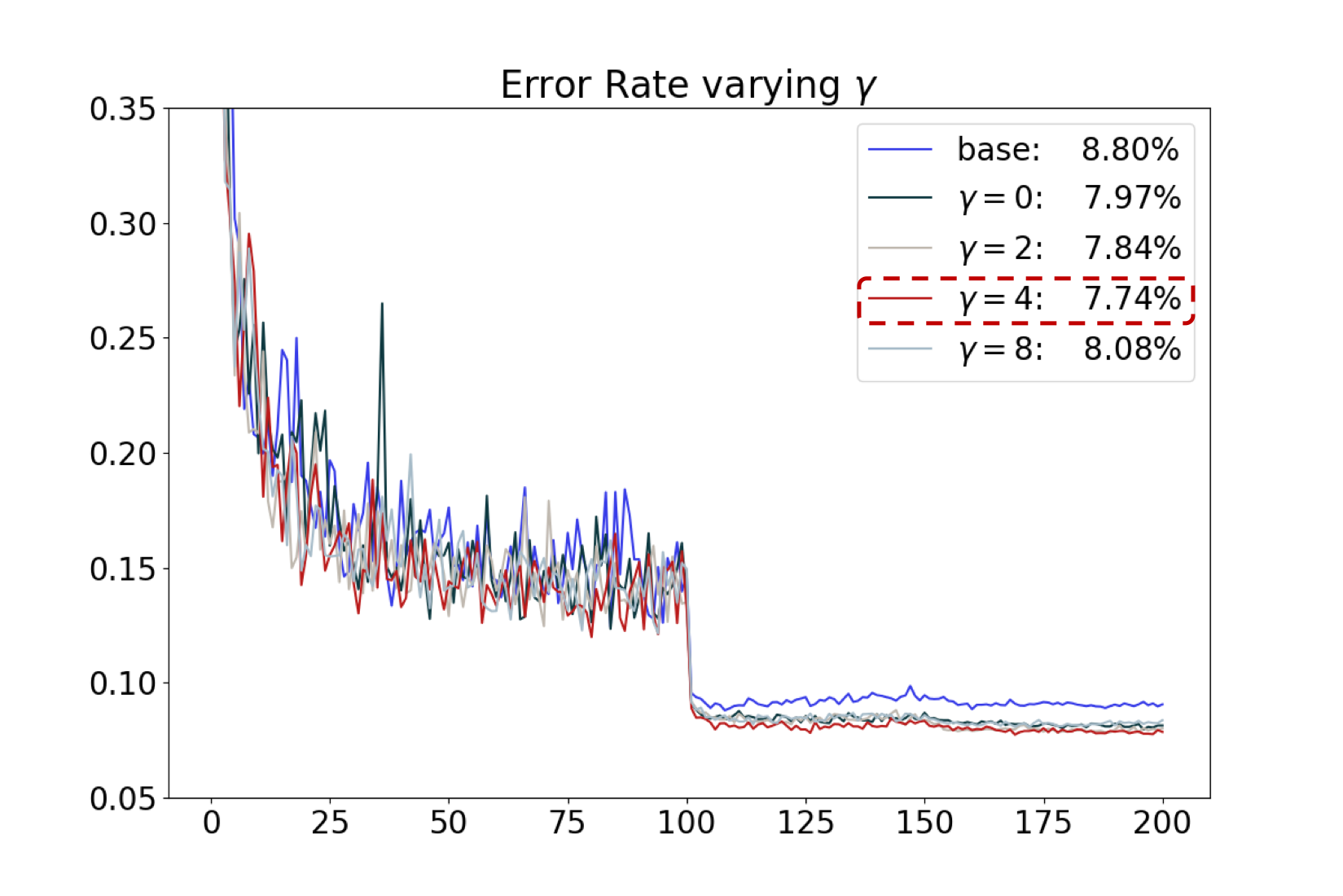}  &
  \includegraphics[width=0.23\textwidth, trim={2.5cm 1cm 2.5cm 1cm}, clip]{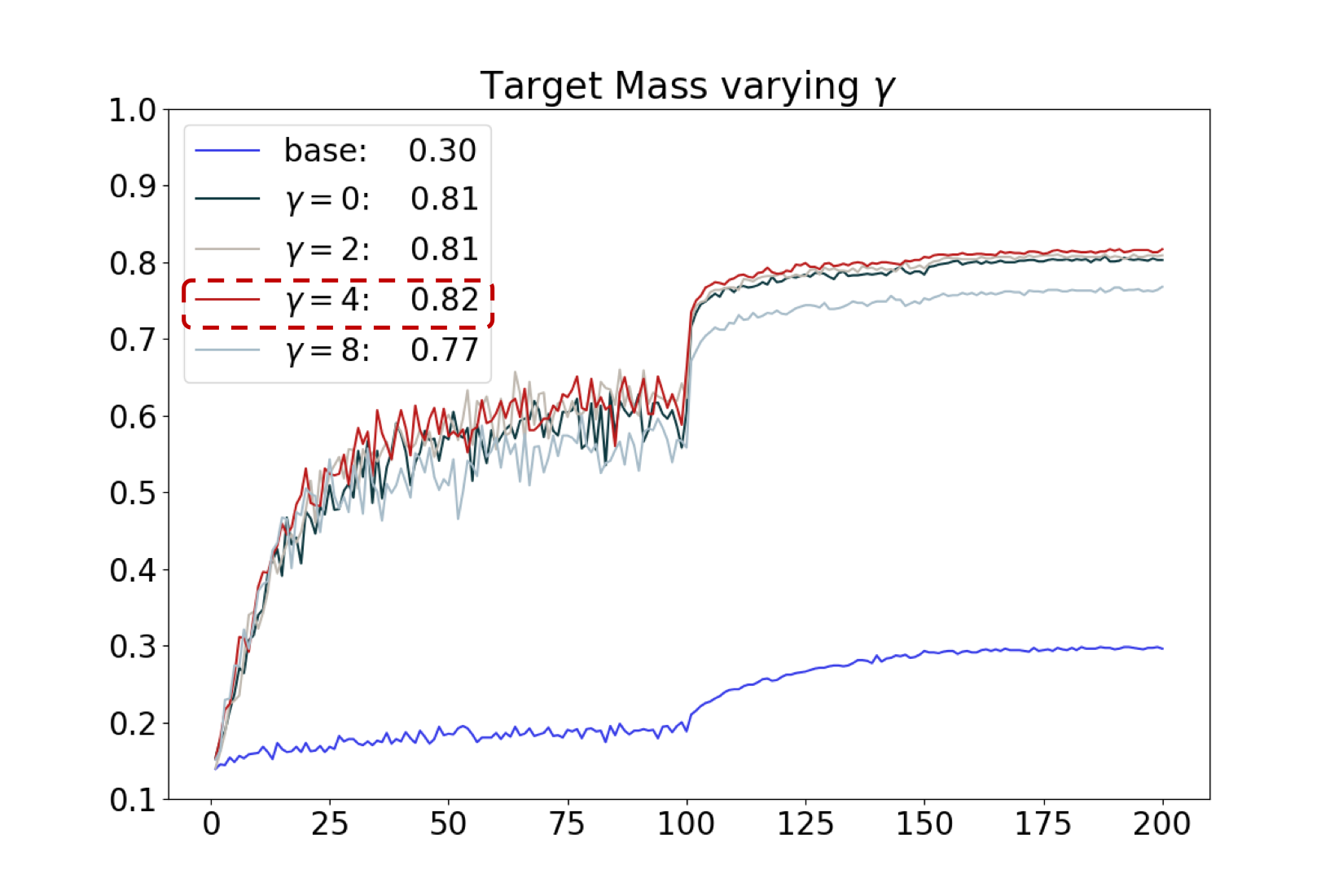} &
  \includegraphics[width=0.23\textwidth, trim={2.5cm 1cm 2.5cm 1cm}, clip]{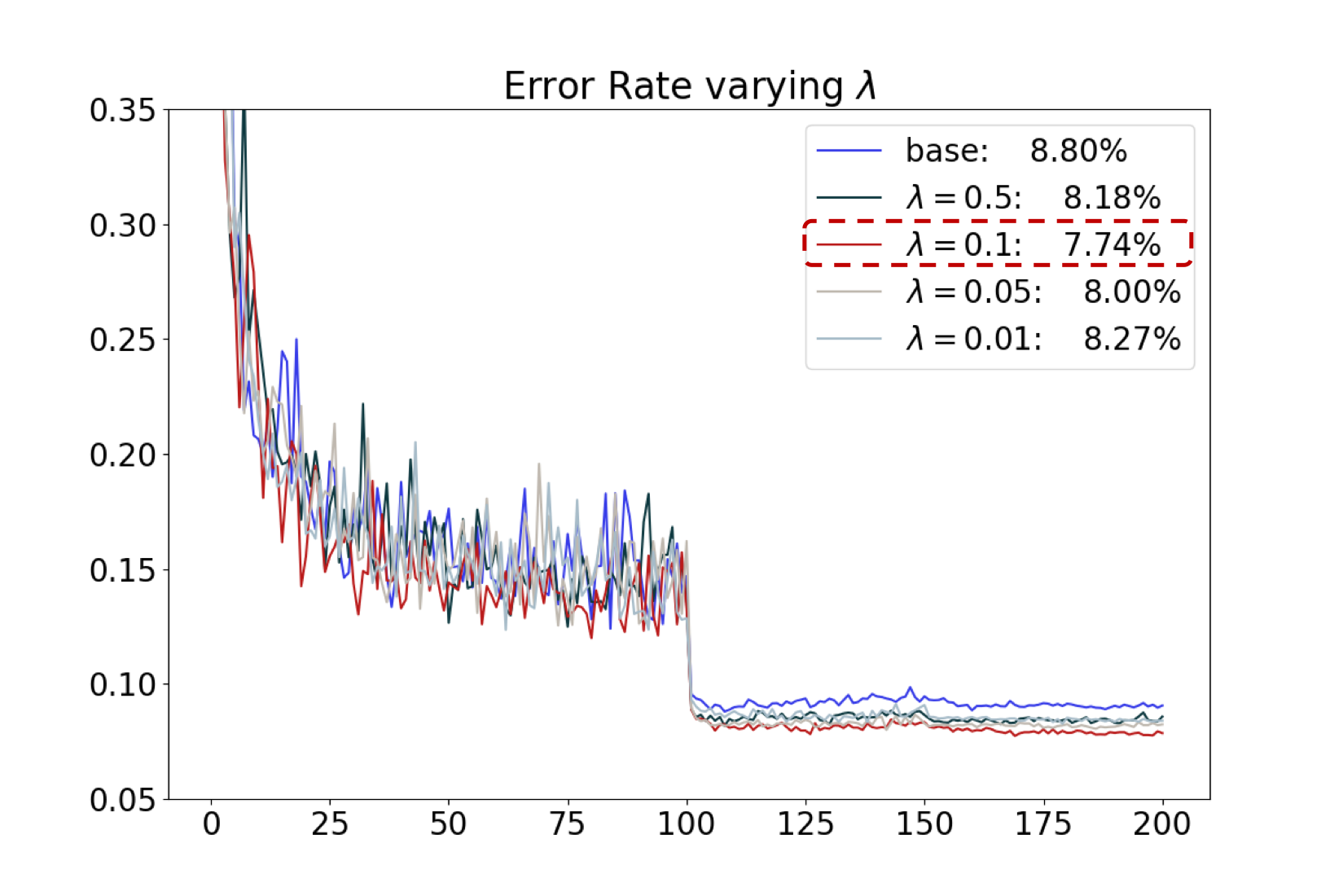} &
  \includegraphics[width=0.23\textwidth, trim={2.5cm 1cm 2.5cm 1cm}, clip]{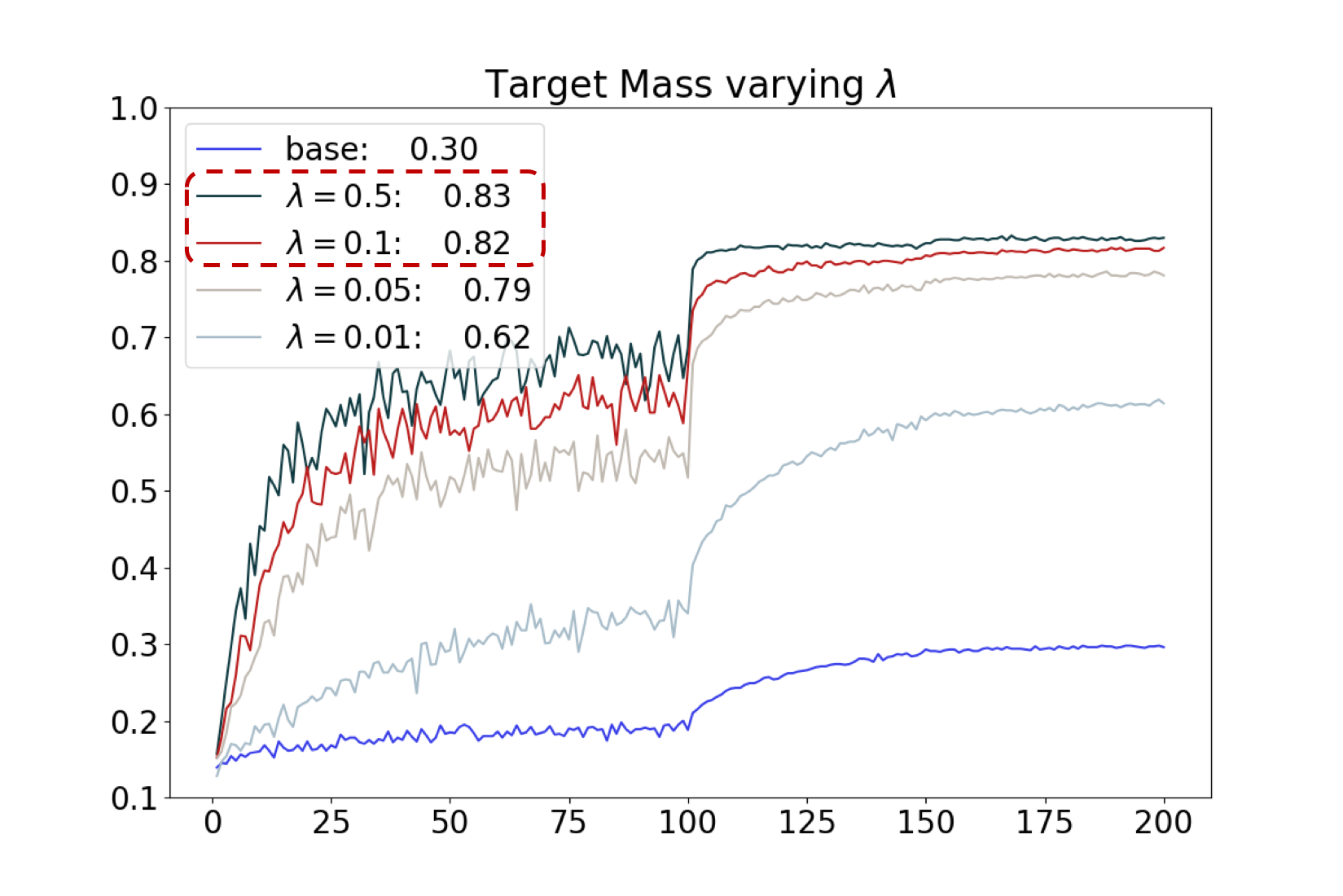} \\
    \end{tabular}
    \captionof{figure}{An ablation study on mini-batch affinity supervision, with the evaluation metric on a test set over epochs (horizontal axis), with the best result highlighted with a red dashed box. Left Plots: classification error rates and target mass with varying focal loss' $\gamma$ parameter. Right Plots: error rates and target mass with varying loss balancing factor $\lambda$ (defined in section \ref{sec:mass_optimization}). 
    }
    \label{fig:batch_aff_analysis}
\end{table*}

\begin{table*}[!ht]
    \begin{minipage}[ht]{.48\textwidth}
        \centering
        %\hspace{-0.3cm}
        \begin{tabular}{c c}
        \includegraphics[width=0.46\textwidth, trim={2cm 1.2cm 2cm 1.2cm}, clip, frame={\fboxrule} {-\fboxrule}]{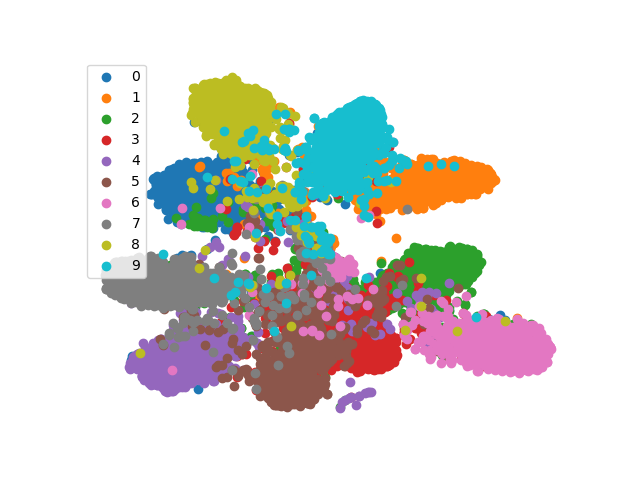} &
        \hspace{-0.25cm}
        \includegraphics[width=0.46\textwidth, trim={2cm 1.2cm 2cm 1.2cm}, clip, frame={\fboxrule} {-\fboxrule}]{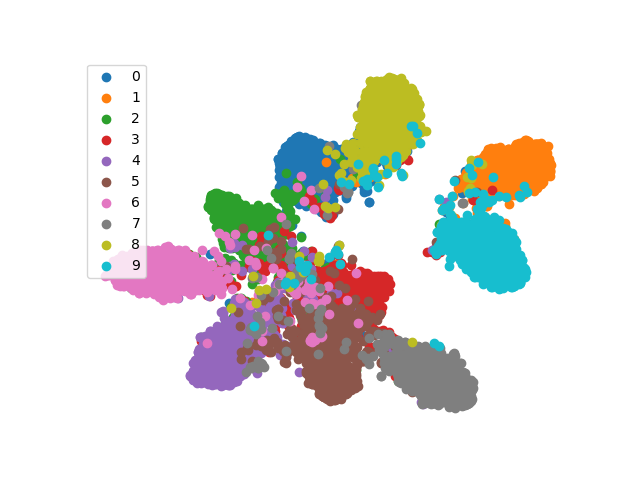}
        \end{tabular}
        %\vspace{-0.5cm}
        \captionof{figure}{Left: t-SNE plot of learned feature representations for a baseline ResNet20 network on CIFAR10 dataset. Right: t-SNE plot for affinity supervised ResNet20 network.
        }
        \label{fig:batch_aff_tsne}
    \end{minipage}
    \hspace{0.25cm}
    \begin{minipage}[ht]{.48\textwidth}
        \centering
         \scalebox{0.85}{
        \begin{tabular}{@{}l|c|c|c}
    		\hline
    		\small
    		CIFAR-10 & ResNet 20 & ResNet 56 & ResNet 110 \\ \hline
    		base CNN & 91.34 $\pm$ 0.27  & 92.24 $\pm$ 0.48  & 92.64 $\pm$ 0.59 \\
    		Affinity Sup & 92.03 $\pm$ 0.21  &  92.90 $\pm$ 0.35  & 93.42 $\pm$ 0.38 \\
    		\hline
    		\hline
    		CIFAR-100 & ResNet 20 & ResNet 56 & ResNet 110 \\ \hline
    		base CNN & 66.51 $\pm$ 0.46 & 68.36 $\pm$ 0.68 & 69.12 $\pm$ 0.63 \\
    		Affinity Sup & 67.27 $\pm$ 0.31  & \textbf{69.79 $\pm$ 0.59} & \textbf{70.5 $\pm$ 0.60}  \\
    		\hline
    		\hline
    		Tiny Imagenet & ResNet 18 & ResNet 50 & ResNet 101 \\ \hline
    		base CNN & 48.35 $\pm$ 0.27 & 49.86 $\pm$ 0.80 &  50.72 $\pm$ 0.82 \\
    		Affinity Sup & \textbf{49.30 $\pm$ 0.21} & \textbf{51.04 $\pm$ 0.68} & \textbf{51.82 $\pm$ 0.71} \\ 
    		\hline
    	\end{tabular}
    	}
    	\caption{Batch Affinity Supervision results. Numbers are classification accuracy in percentages. CIFAR results are reported over 10 runs and tiny ImageNet over 5 runs.}
    	\label{tab:batch_aff_cifar}
    \end{minipage}
\end{table*}
\subsection{Scene Categorization Task}
\label{sec:scenecategorization}
For scene categorization we adopt the base visual attention network (Figure \ref{fig:net_architecture}, part A), and add an additional scene task branch (Figure \ref{fig:net_architecture}, part C) to fine tune it on MIT67, as discussed in Section \ref{sec:scene_net}. Table \ref{tab:scenecategorization} shows the results of applying this model to the MIT67 dataset.
We refer to the baseline CNN as ``CNN'' (first column), which is an ImageNet pretrained ResNet101 model directly applied to an image classification task. In the second column, we first acquire a COCO pretrained visual attention network (Figure \ref{fig:net_architecture}, part A), and fine tune it using only the scene level feature $F_S$ (Figure \ref{fig:net_architecture}, part C). 
In the third column, for the same COCO pretrained visual attention network, we concatenate object proposals' ROI pooling features with $F_S$ to serve as meta scene level descriptor. In the fourth and fifth columns, we apply the full scene architecture in Figure \ref{fig:net_architecture} part C, but with a visual attention network that is pretrained without and with (supervised) target affinity loss, respectively. The affinity supervised case (fifth column) demonstrates a non-trivial improvement over the baseline (first to third columns) and also significantly outperforms the unsupervised case (fourth column). This demonstrates that the attention weights learned solely by minimizing detection loss do not generalize well to a scene task, whereas those learned by affinity supervision can. 

\subsection{Mini-Batch Affinity Supervision}
We conducted model ablation study on $\gamma$ and $\lambda$ parameters, introduced in section \ref{sec:aff_sup}, as summarized in Figure \ref{fig:batch_aff_analysis}. In the subsequent experiments we chose $\gamma = 4$ and $\lambda = 0.1$ based on the ablation plots for error rates in Figure \ref{fig:batch_aff_analysis}. 
%\vspace{-0.5cm}
\paragraph{Convergence of Target Mass.}
We plot results showing convergence of the target affinity mass during learning in Figure \ref{fig:batch_aff_analysis}. There is a drastic improvement over the baseline target mass convergence, when affinity supervision is enabled. The chosen $\lambda = 0.1$ empirically gives sufficient convergence on Target mass (right-most in Figure \ref{fig:batch_aff_analysis}). 
%In addition, the parameters of choice, being $\gamma = 4$, also gives the best converged Target mass (left-mid in Figure \ref{fig:batch_aff_analysis}). 
%\vspace{-0.5cm}
\paragraph{Per-class feature separation.}
A comparison of t-SNE \cite{maaten2008visualizing} plots on learned feature representations from 1) baseline CNN and 2) CNN supervised with affinity mass loss is presented in Figure \ref{fig:batch_aff_tsne}. Note that the feature separation between different classes is better in our case. 
%\vspace{-0.5cm}
\paragraph{Results.} We now summarize the results for mini-batch affinity learning on CIFAR10, CIFAR100 and TinyImageNet in Table \ref{tab:batch_aff_cifar}. 
Overall, we have a consistent improvement over the baseline, when using the affinity supervision in mini-batch training. In particular, for datasets with a large number of categories, such as CIFAR100 (100-classes) and tiny ImageNet (200-classes), the performance gain is above 1\%. Another advantage of affinity supervision is that we do not introduce any additional network layers or parameters, except for the need for computing the $N \times N$ affinity matrix and its loss. Therefore, the we found the training run-time of affinity supervision very close to the baseline CNN.
\section{Conclusion}
\label{sec:conclusion}
 In this paper we have addressed an overlooked problem in the computer vision literature, which is the direct supervision of affinity graphs applied in deep models. Our main methodological contribution is the introduction of a novel target affinity mass, and its optimization using an affinity mass loss. These novel components lead to demonstrable improvements in relationship retrieval. In turn, we have shown that the improved recovery of relationships between objects boosts scene categorization performance. We have also explored a more general problem, which is the supervision of affinity in mini-batches. Here, in diverse visual recognition problems, we see improvements once again. Given that our affinity supervision approach introduces no additional parameters or layers in the neural network, it adds little computational overhead to the baseline architecture. Hence it can be adopted by the community for affinity based training in other computer vision applications as well.
 
\section*{Acknowledgments} We thank the Natural Sciences and Engineering Research Council of Canada (NSERC) and Adobe Research, for research funding.

{\small
\bibliographystyle{ieee_fullname}
\bibliography{egbib}
}

\newpage
\appendix
\paragraph{\Large Supplementary Material}
\section{Runtime Comparisons}
\begin{table}[!ht]
    \centering
    \begin{tabular}{l|cc}
        \toprule
        VOC07-Res101  & 1-epoch Runtime   &  GPU Memory  \\
        \midrule
        Baseline & 79.7 minutes &  3137 MB   \\
        Baseline + $\mathcal{L}_G$ & 82.3 minutes &  3141 MB  \\
        \bottomrule
        \toprule
        CIFAR10-Res110  & 1-epoch  Runtime   & GPU Memory   \\
        \midrule
        Baseline & 34.7 seconds &  2117 MB   \\
        Baseline + $\mathcal{L}_G$ & 35.9 seconds &  2139 MB  \\
        \bottomrule
    \end{tabular}
    \caption{Training time (runtime) versus GPU memory consumption between the baseline and our affinity supervised network (denoted as ``Baseline + $\mathcal{L}_G$''). For VOC07 we apply RelNet \cite{hu2017relation} as the baseline. Its affinity supervised version is discussed in Section 4.2 of our article. For CIFAR-10, the baseline is a ResNet-110 network \cite{he2016deep} and its affinity supervised version is discussed in Section 5.3 of our article.}
    \label{tab:runtime}
\end{table}
We provide efficiency analysis for visual attention networks as well as mini-batch training, with and without the affinity loss. The results are summarized in Table (\ref{tab:runtime}), with a ResNet101 structure trained on VOC07 dataset for attention networks and a ResNet110 structure trained on CIFAR10 for mini-batch training. For all experiments reported in Table (\ref{tab:runtime}), we used a machine configuration of a single Titan XP GPU, an Intel Core i9 CPU and 64GBs of RAM. The affinity supervision did introduce a small percentage increment to runtime and GPU memory, but the benefits are appreciable, as reflected in the multiple experiments reported in our article.

\begin{table*}[t]
    \centering
    \begin{tabular}{c c | c c}
    %\hline
    \multicolumn{2}{c|}{Cases with improved results} & \multicolumn{2}{c}{Cases with similar results}  \\ 
    \hline
    RelNet \cite{hu2017relation} & with Affinity-Sup & RelNet \cite{hu2017relation} & with Affinity-Sup \\
    \centering
      \includegraphics[width=0.23\textwidth, trim={0 1.3cm 0 4cm}, clip]{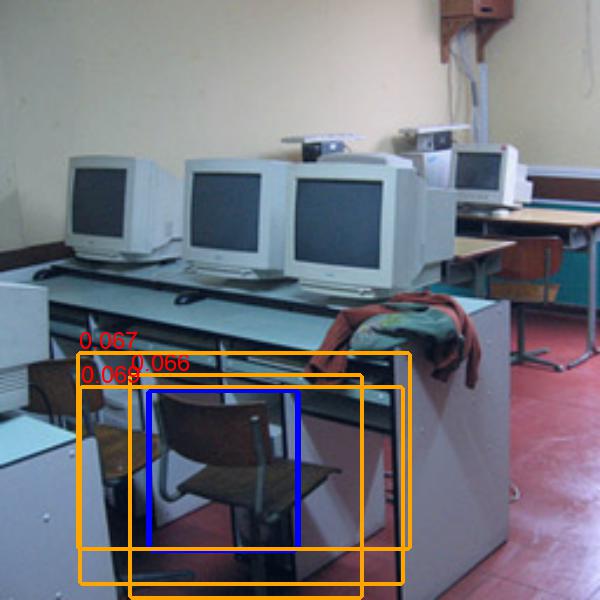} & 
      \includegraphics[width=0.23\textwidth, trim={0 1.3cm 0 4cm}, clip]{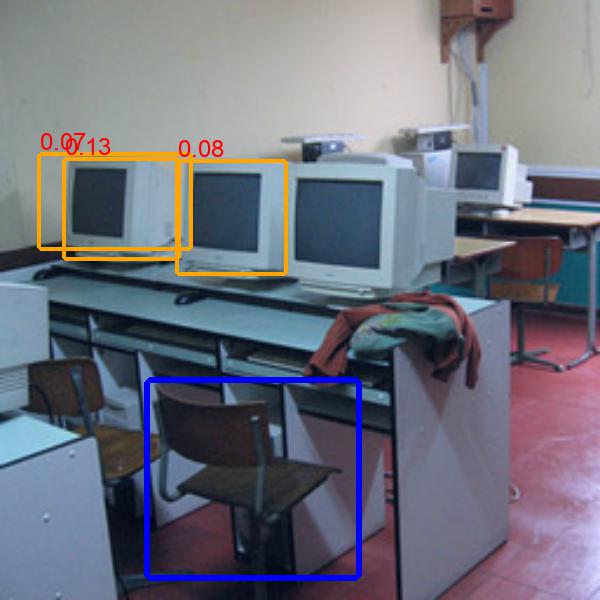} &
      \includegraphics[width=0.23\textwidth, trim={0 0cm 0 0cm}, clip]{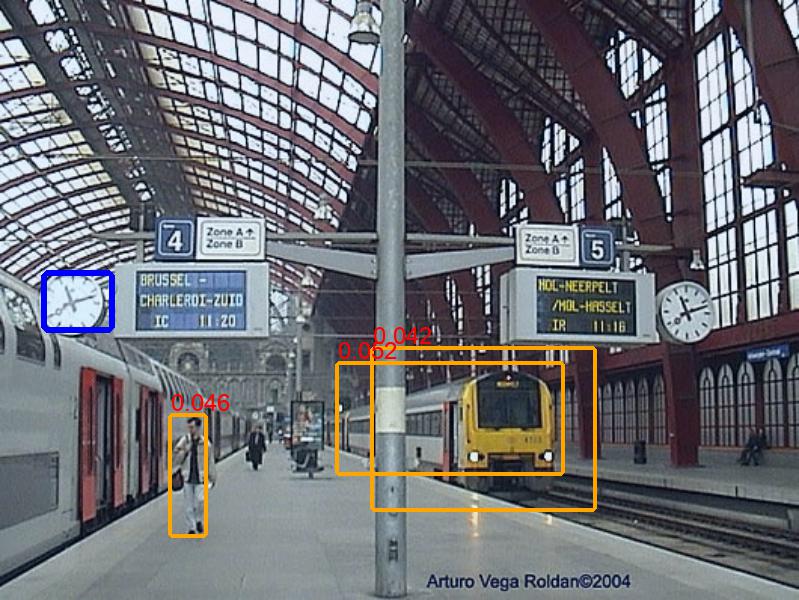} & 
      \includegraphics[width=0.23\textwidth, trim={0 0cm 0 0cm}, clip]{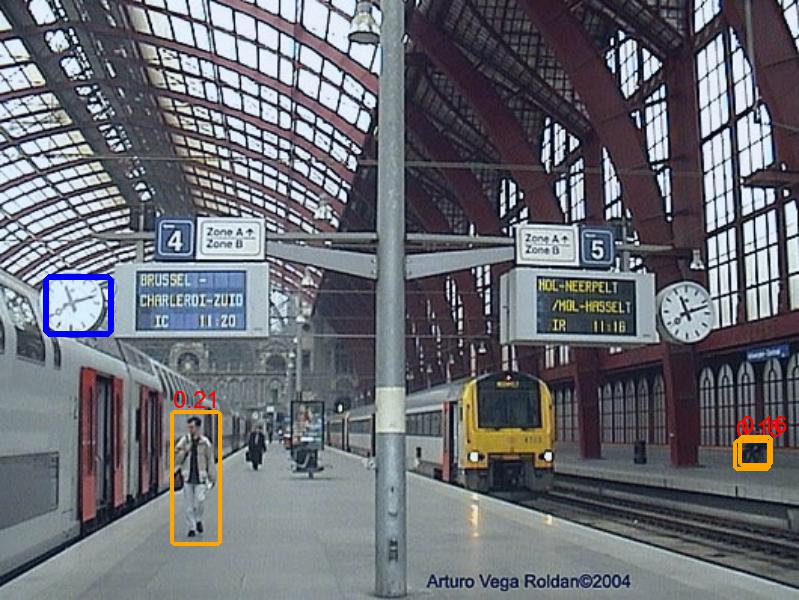}  \\
      
      \includegraphics[width=0.23\textwidth, trim={0 0cm 3.5cm 0cm}, clip]{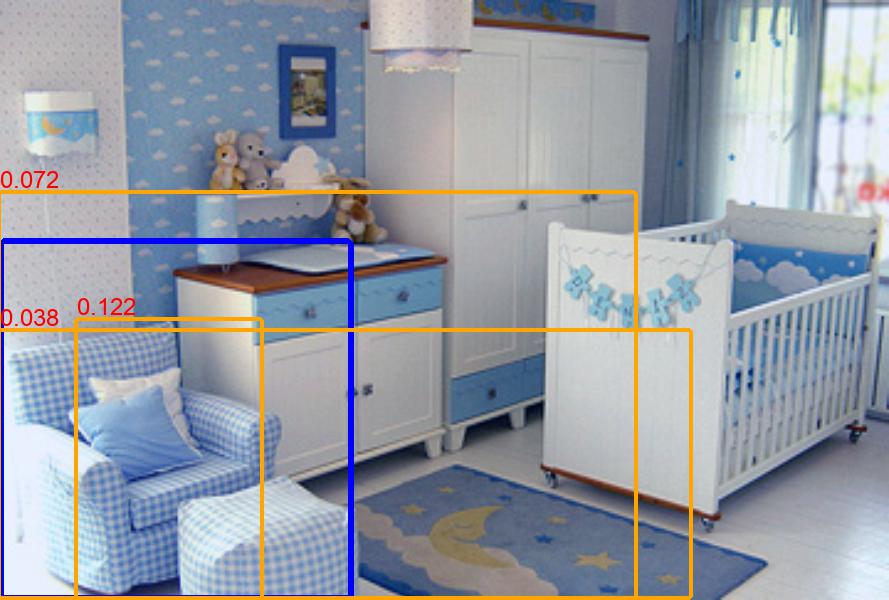} & 
      \includegraphics[width=0.23\textwidth, trim={0 0cm 3.5cm 0cm}, clip]{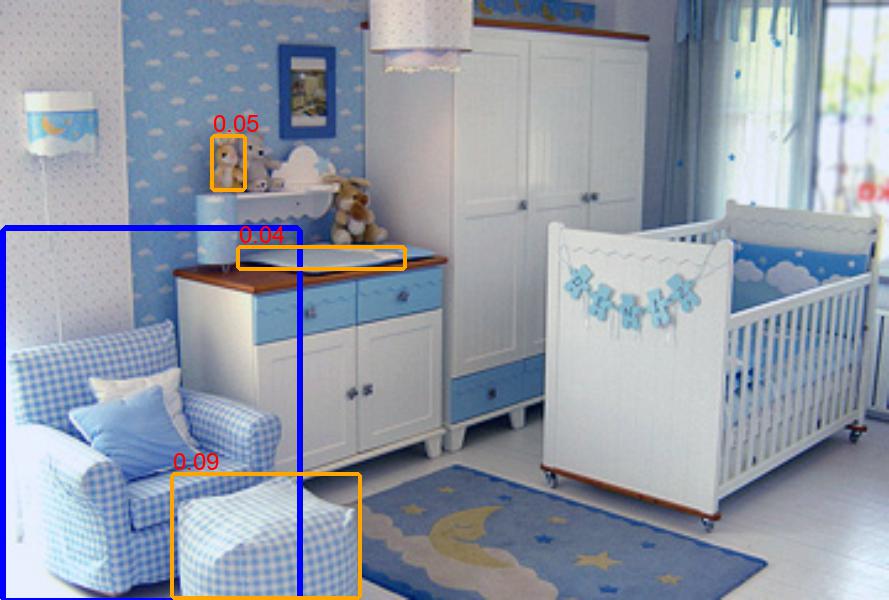} &
      \includegraphics[width=0.23\textwidth, trim={0 0cm 4cm 0cm}, clip]{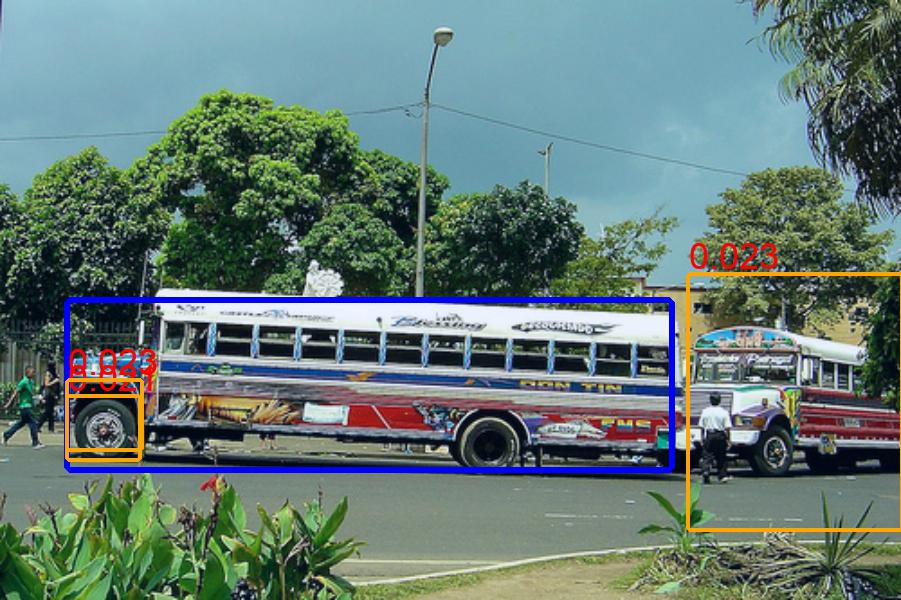} & 
      \includegraphics[width=0.23\textwidth, trim={0 0cm 4cm 0cm}, clip]{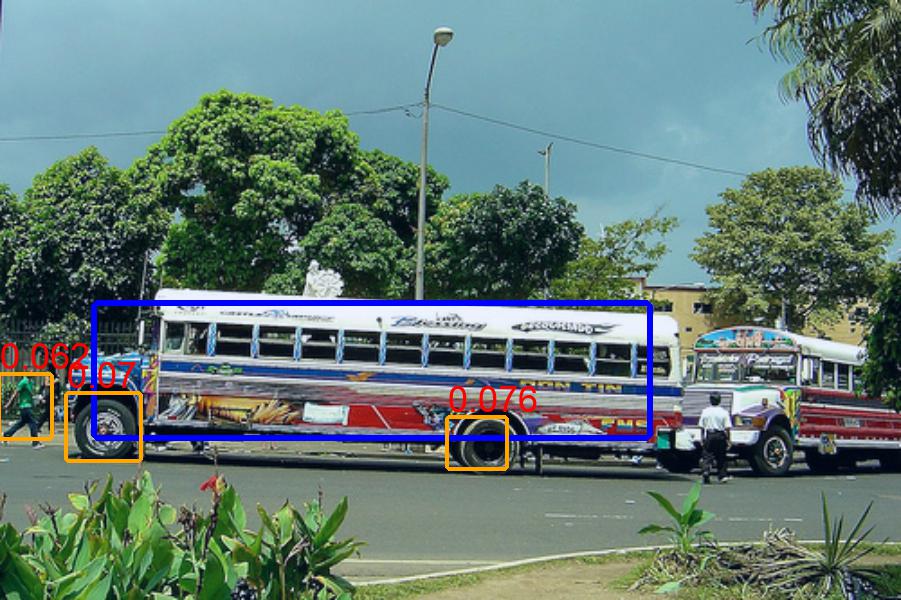} 
      \\
      \includegraphics[width=0.23\textwidth, trim={0 3cm 0 1.5cm}, clip]{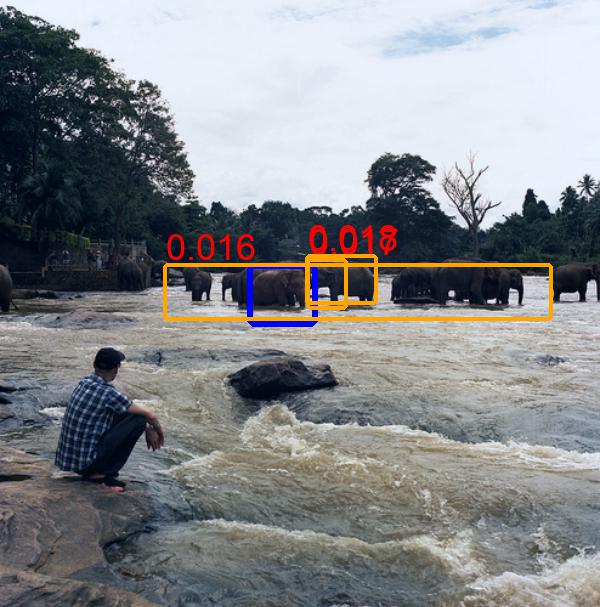} & 
      \includegraphics[width=0.23\textwidth, trim={0 3cm 0 1.5cm}, clip]{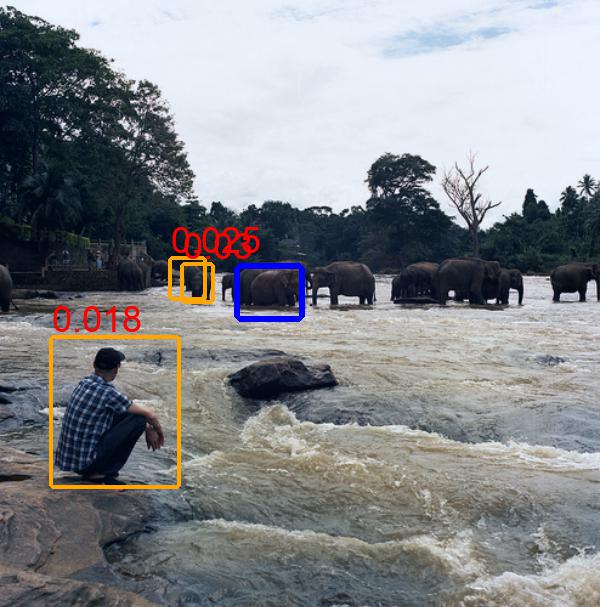} &
      \includegraphics[width=0.23\textwidth, trim={0 0cm 5cm 0cm}, clip]{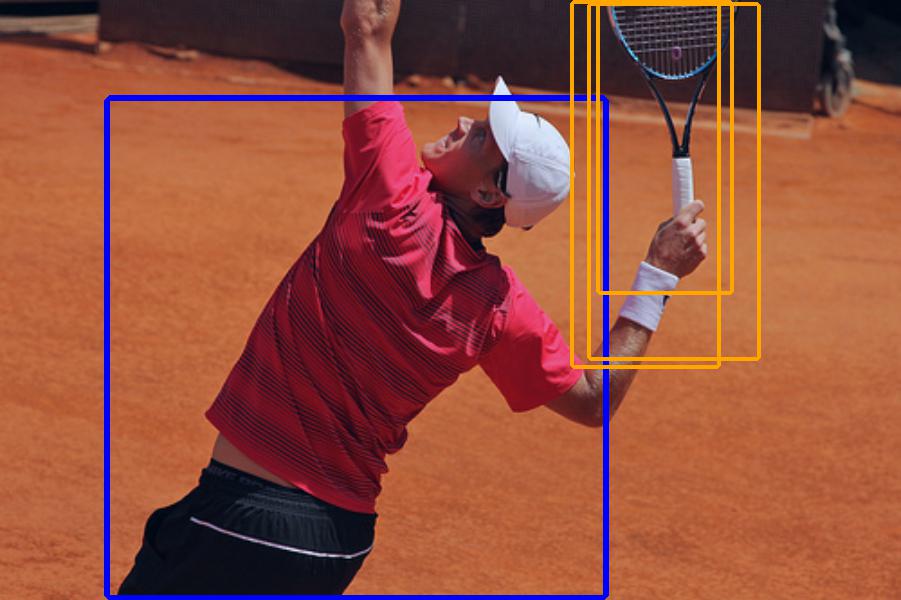} & 
      \includegraphics[width=0.23\textwidth, trim={0 0cm 5cm 0cm}, clip]{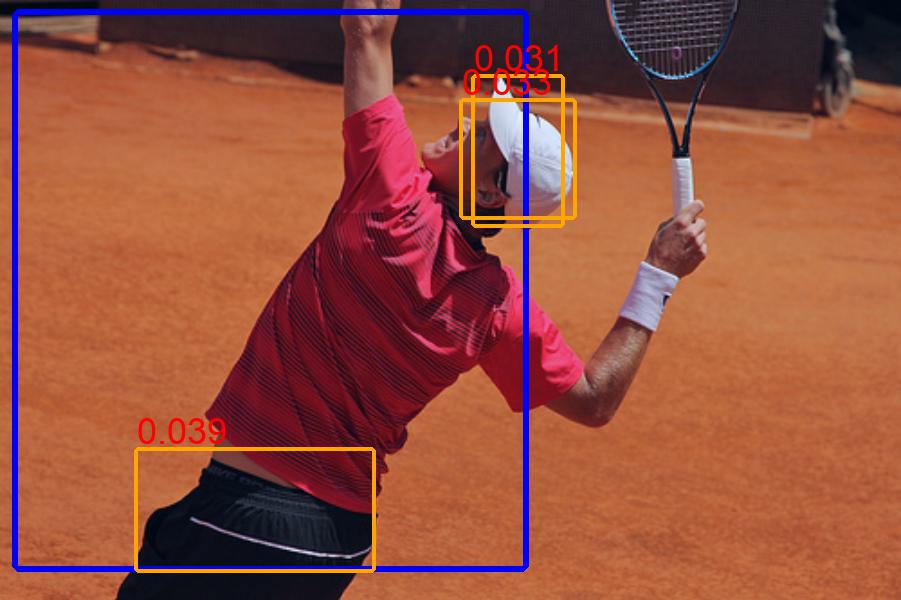} 
      \\
      \includegraphics[width=0.23\textwidth, trim={0 1cm 3cm 0cm}, clip]{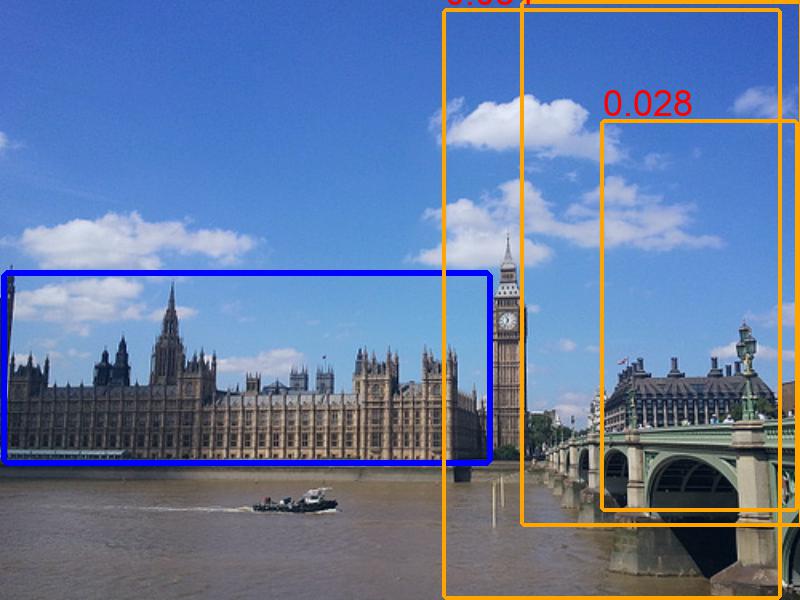} & 
      \includegraphics[width=0.23\textwidth, trim={0 1cm 3cm 0cm}, clip]{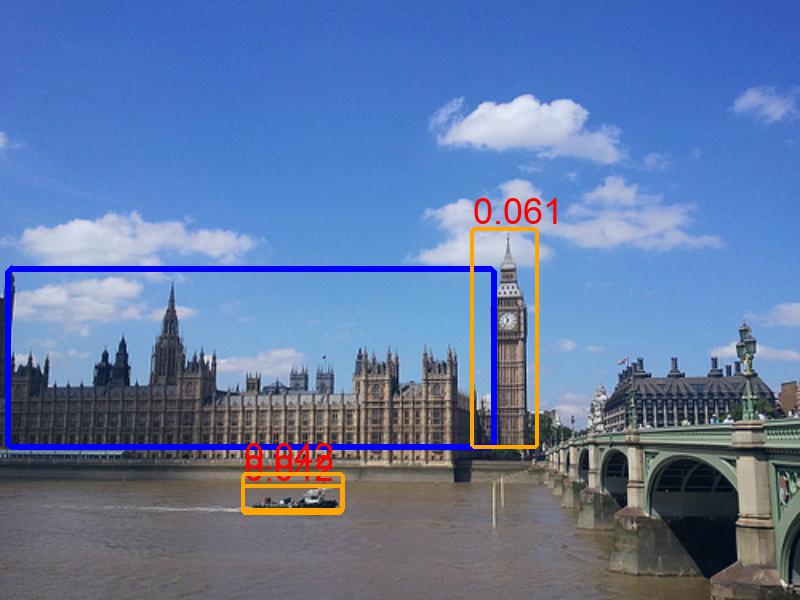} &
      \includegraphics[width=0.23\textwidth, trim={2cm 0cm 3cm 0cm}, clip]{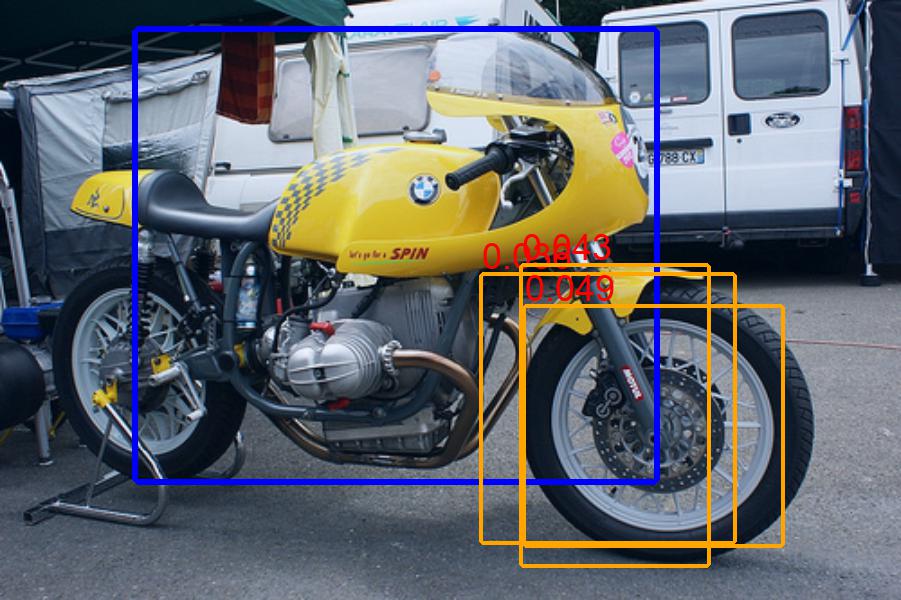} & 
      \includegraphics[width=0.23\textwidth, trim={2cm 0cm 3cm 0cm}, clip]{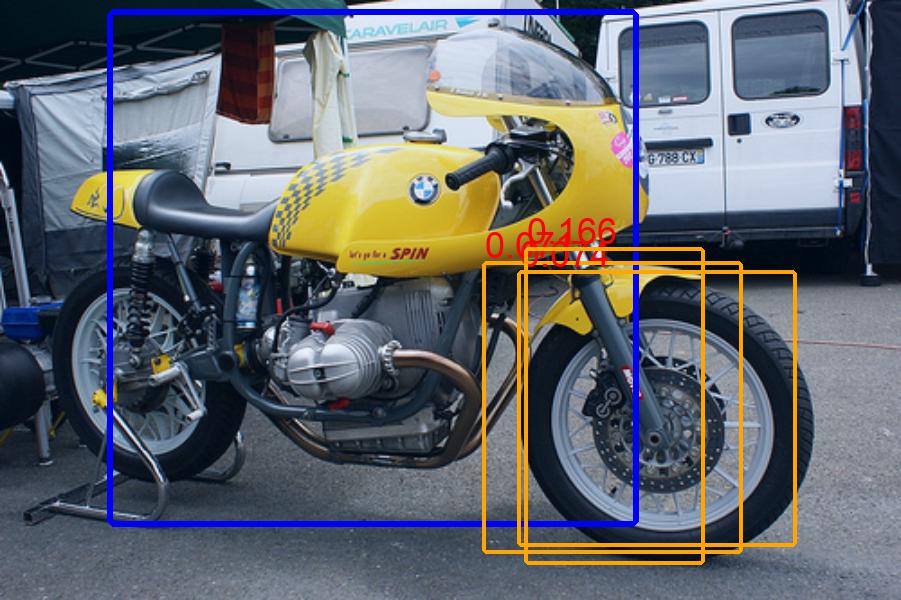} 
      \\
      
      \includegraphics[width=0.23\textwidth, trim={0 2.5cm 0 2cm}, clip]{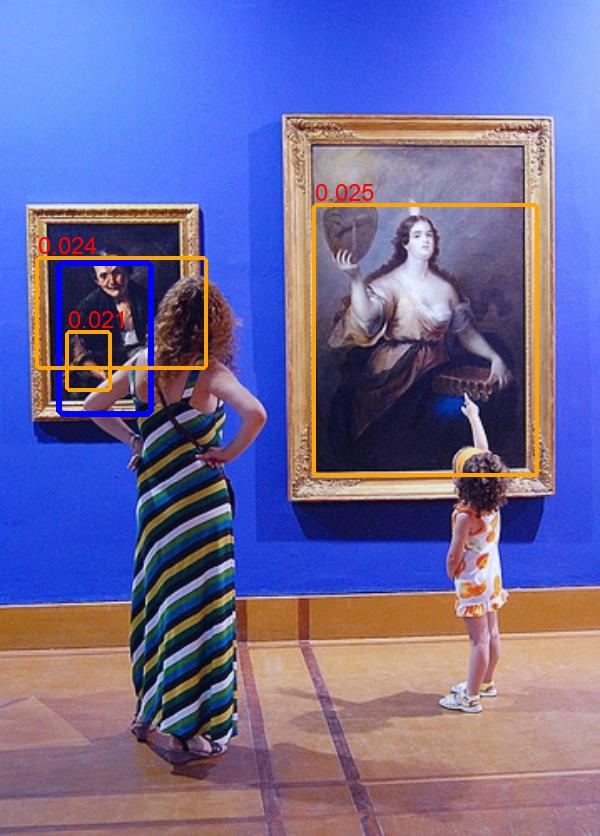} & 
      \includegraphics[width=0.23\textwidth, trim={0 2.5cm 0 2cm}, clip]{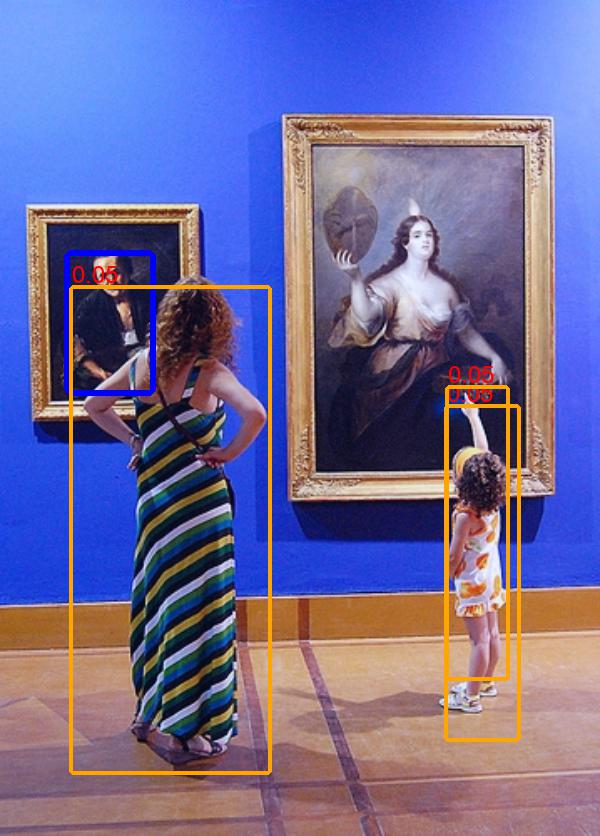} &
      \includegraphics[width=0.23\textwidth, trim={0 2cm 0 2cm}, clip]{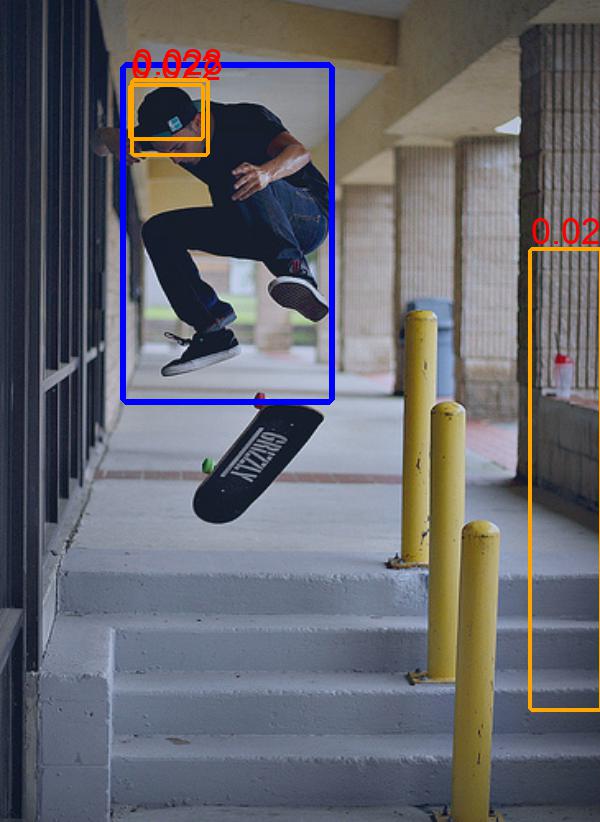} & 
      \includegraphics[width=0.23\textwidth, trim={0 2cm 0 2cm}, clip]{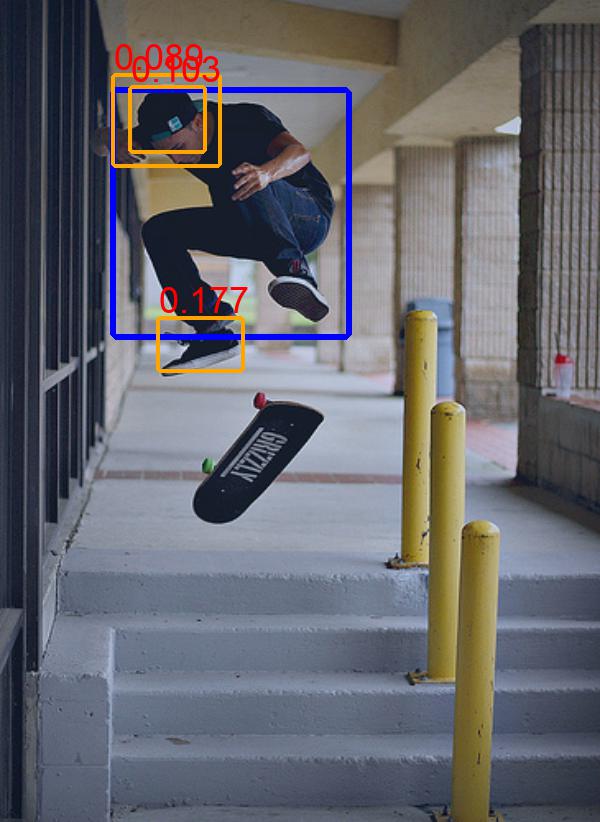} 
      \\
    \end{tabular}
    \captionof{figure}{Additional comparisons of recovered relationships on test images, including cases with a clear improvement over the baseline (left) and cases where the results are comparable). The affinity supervision applied to acquire these results do not use human annotated relationship labels. See the text in Figure 1 of the main article for details about the representation. Zoom in on the PDF file to see the attention weight values.
    }

    \label{fig:visual_rel_supp}
\end{table*}

\section{Object Detection Results}
As promised in our article, we present the object detection results of affinity supervised attention networks in Table \ref{tab:objectdetection}. We report results on VOC07 and the COCO split we applied in our article.
In both cases we improve upon the baseline and slightly outperform the unsupervised case (similar to Relation Networks \cite{hu2017relation}). This suggests that relation weights learned using affinity supervision are at least as good as those from the unsupervised case \cite{hu2017relation}, in terms of object detection performance.
\begin{table}[h]
	\centering
	\scalebox{0.9}{
	\begin{tabular}{@{}l|c|c|c}
		\hline
		VOC07 & base F-RCNN & RelNet  \cite{hu2017relation} & RelNet + $\mathcal{L}_{G}$ \\ \hline
		avg mAP (\%) & 47.0  & 47.7 $\pm$ 0.1  & \textbf{48.2 $\pm$ 0.1}   \\
		mAP@0.5 (\%) & 78.2  & 79.3 $\pm$ 0.2  & \textbf{79.9 $\pm$ 0.2} \\
		\hline
		\hline
		mini COCO & base F-RCNN & RelNet  \cite{hu2017relation} & RelNet + $\mathcal{L}_{G}$ \\ \hline
		avg mAP (\%) & 26.8  & 27.5  & \textbf{27.9}    \\
		mAP@0.5 (\%) & 46.6  & 47.4  & \textbf{47.8}    \\
		\hline
	\end{tabular}
	}
	\caption{Object Detection Results. mAP@0.5: average precision over a bounding box overlap threshold set to $IOU=0.5$. avg mAP: averaged mAP over multiple bounding box overlap thresholds. VOC07 experiments are reported over 3 runs, demonstrating stability. $\mathcal{L}_{det}$ stands for detection task loss as defined in \cite{fasterRCNN} and $\mathcal{L}_{G}$ for the target affinity mass loss defined in Section 3.3 of the main article. }
	\label{tab:objectdetection}
\end{table}

\paragraph{Convergence of Target Mass}
We also provide results showing the convergence of target mass in the context of visual attention networks, in Table (\ref{tab:target_mass_attn}). It is evident that the affinity mass loss succeeds in optimizing the target mass, when compared with a baseline Relation Network model \cite{hu2017relation}. This is also indirectly supported by the dramatic improvement in the recall metric, reported in Table 1 of the main article and Table (\ref{tab:mass_def_ablation}), when comparing between the baseline visual attention network and its affinity supervised version.

\begin{table}[h]
    \centering
    \begin{tabular}{lcc}
        \toprule
        COCO $\mathcal{M}$ &  Training   &  Testing  \\
        \midrule
        RelNet \cite{hu2017relation} & 0.020 &  0.013   \\
        RelNet + $\mathcal{L}_G$  & 0.747 &  0.459   \\
        \bottomrule
    \end{tabular}
    \caption{We compare target mass values for a visual attention network supervised with (RelNet + $\mathcal{L}_G$) and without (RelNet) the affinity mass loss, using the target constructed in Section 4.1 of our main article. 
    The values reported are evaluated on the COCO split, that is described in the experiment section.}
    \label{tab:target_mass_attn}
\end{table}

\section{Additional Illustrations}
Additional examples of visual relationships, recovered using baseline Relation Networks \cite{hu2017relation} and its affinity supervised version (discussed in Section 4.2 of our article), are provided in Figure \ref{fig:visual_rel_supp}. Here we allow all regions with different object class labels to have a potential relation with one another. No human annotated relationship labels are used during training. We include both examples showing improvement and ones where the results are comparable. 
% In most scenarios, the added affinity supervision demonstrated improvement, or at least being comparable, on relationship recovery task when compared to a baseline visual attention network. 
Whereas the baseline method can be effective at times (Figure (\ref{fig:visual_rel_supp}) right half), the affinity supervision improves its consistency. This claim is also supported by the relationship proposal results on Visual Genome reported in Section 6.5 of the main article.

\begin{figure*}[ht]
    % \left
    \centering
    \includegraphics[width=\textwidth, trim={0cm 0cm 0cm 0cm}, clip]{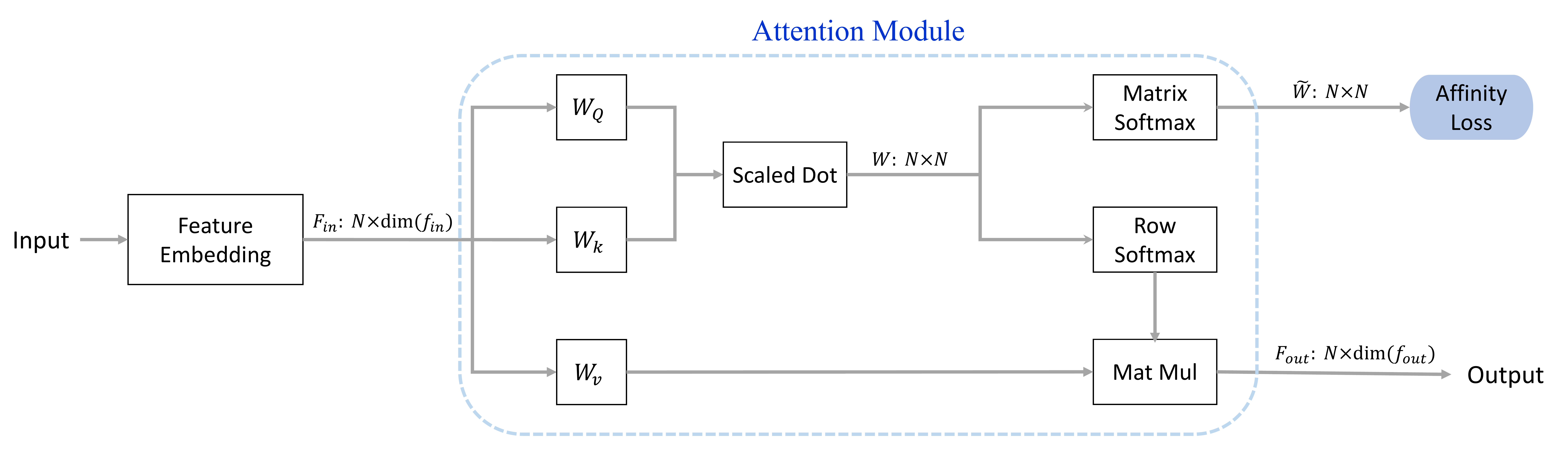}
    \caption{A detailed illustration of the structure inside the ``attention module'' shown in Figure 2 of the main article. An explanation of each step in this figure is provided at the beginning of Section 4 in the article.
    }
    \label{fig:structure_attn_module}
\end{figure*}
\section{Structure inside Attention Module}
A structural overview of the visual attention module used throughout the main article is presented in Figure (\ref{fig:structure_attn_module}). 

\section{Supervision Targets} 
\label{sec:flexible_target}
The proposed loss requires a task specific design of a supervision target $\mathcal{T}$. The flexibility in choosing this target is one of the core advantages of our affinity supervision, that is, it can be any user designed matrix target. In fact, in the experiments reported in the main article, we construct this matrix automatically using only object class labels, i.e., no labeled relations are required. In the context of mini-batch training, the target design is straightforward. Here we aim to reduce the within-class feature distances between batch images. Thus, a same-category target $\mathcal{T}$ is adopted. This target increases the similarity metric between same-class connections in the weight matrix $\Tilde{\mathcal{W}}$ and because of the matrix-wise softmax activation, connections between instances from different classes are suppressed.

For the case of visual attention networks, various supervision targets can be applied to adapt the method for different downstream applications. In the main draft, our goal is to improve visual recognition using contextual features aggregated by the attention module, with improved object-wise relationship recovery. Thus, the supervision target emphasizes relationships between instances from different categories. However, given a distinct vision task, such as learning human-to-human interaction or human-to-X interaction, the target $\mathcal{T}$ could also be constructed by only selecting human-to-human or human-to-X instance connections, while suppressing other possibilities.

To support the idea that the target $\mathcal{T}$ is adaptive, we provide an exemplar ablation study on VOC07 detection task. We first consider the supervision target proposed in the main draft as \textbf{different-category supervision}. We now consider the case where attention between distinct objects belonging to the same category is {\em also} of interest, leading to same-category connections in the target matrix $\mathcal{T}$. %That is, as long as object proposal $a$ and object proposal $b$ are different object instances in an image, we consider a possible relationship between them and assign $t^{ab} = \mathcal{T}[a,b] = 1$.
We refer to this as \textbf{different-instance supervision}. We provide a visual example of the above mentioned supervision targets in Figure \ref{fig:visual_target}. In Table \ref{tab:sup_target}, we provide object detection results on VOC07 when supervising the affinity graph using different targets.
\begin{table}[h]
    \centering
    \begin{tabular}{c | c}
    %\hline
    Different-Category & Different-Instance \\
    \toprule
    \centering
      \includegraphics[width=0.22\textwidth, trim={0 0cm 0 0cm}, clip]{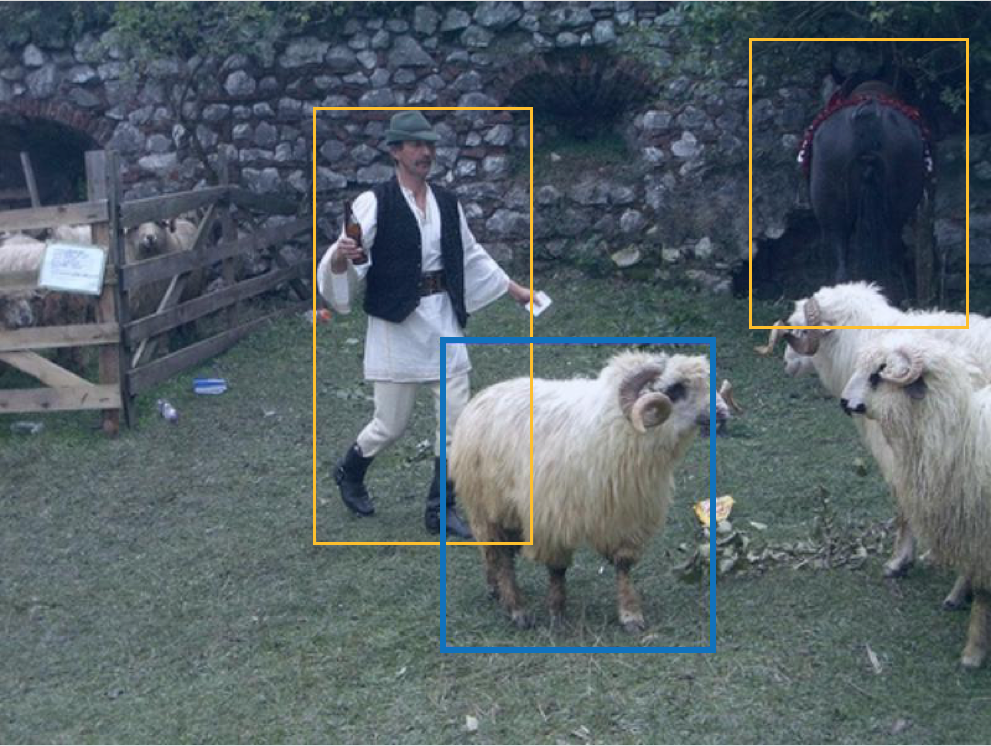} & 
      \includegraphics[width=0.22\textwidth, trim={0 0cm 0 0cm}, clip]{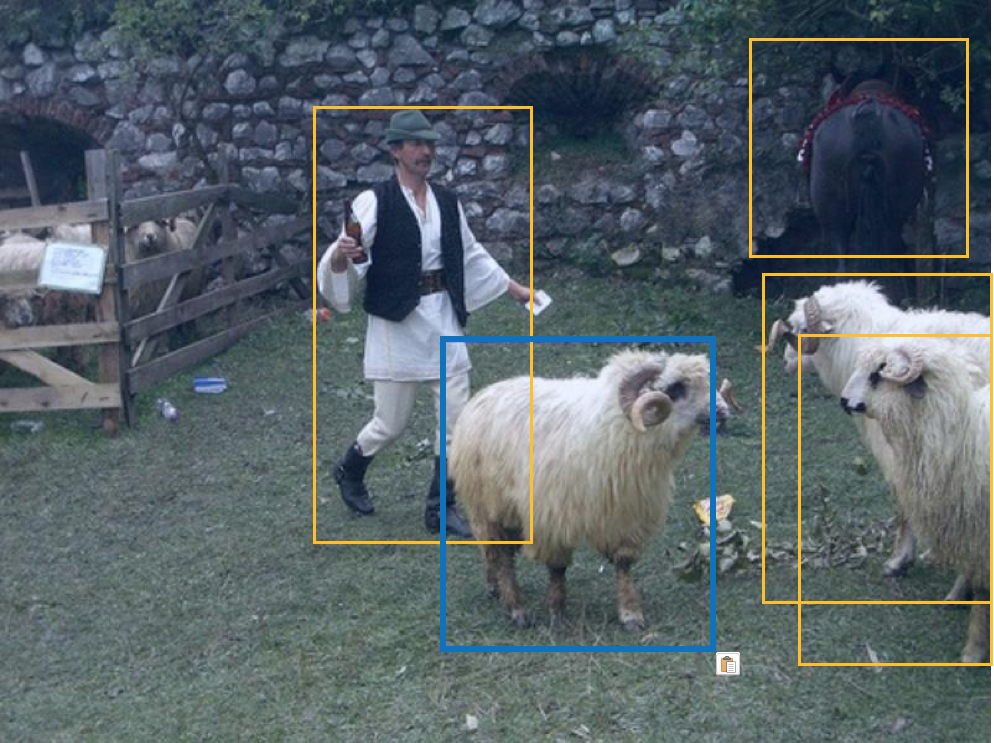} \\
    \end{tabular}
    \captionof{figure}{The visualization of supervision targets for attention networks. The blue box indicates a fixed reference object $a$ and the orange boxes indicate the objects $b$ that have a ground truth relationship with $a$, for which we assign $\mathcal{T}[a,b] = 1$. Left: different category supervision. Note that the sheep in the blue box is \textit{not} related to the other sheep in the image. Right: different instance supervision.% The sheep in the blue box now has a relationship to other sheep (in yellow boxes). \cw{add red lines between gt connections.}
    }
    \label{fig:visual_target}
\end{table}

\begin{table}[!h]
    \centering
    \scalebox{1.0}{
    \begin{tabular}{l|cc}
        \toprule
         VOC07 varying $\mathcal{T}$ & Diff-Instance  & Diff-Category    \\
        \hline
         avg mAP (\%) & 47.6 $\pm$ 0.1 & 48.2 $\pm$ 0.2\\
         mAP@0.5 (\%) & 79.5 $\pm$ 0.2 & 79.9 $\pm$ 0.2\\
        \bottomrule
    \end{tabular}
    % \begin{tabular}{l|ccc}
    %     \toprule
    %      VOC07 & Diff Instance & Same Category*  & Diff Category    \\
    %     \hline
    %      avg mAP (\%) & 47.6 $\pm$ 0.1 & ? & 48.2 $\pm$ 0.2\\
    %      mAP@0.5 (\%) & 79.5 $\pm$ 0.2 & ? & 79.9 $\pm$ 0.2\\
    %     \bottomrule
    % \end{tabular}
    }
    \caption{Detection results on the VOC07 dataset when varying supervision targets, where we show mean accuracy over 3 runs. }
    \label{tab:sup_target}
\end{table}
In summary, the affinity supervision can be adapted to different targets, to achieve various goals or to handle distinct downstream tasks. However, the successful construction of such a target is task dependent.

\begin{table}[ht]
\centering
\begin{tabular}{c}
     \includegraphics[width=0.47\textwidth, trim={2cm 7cm 2cm 8.5cm}, clip]{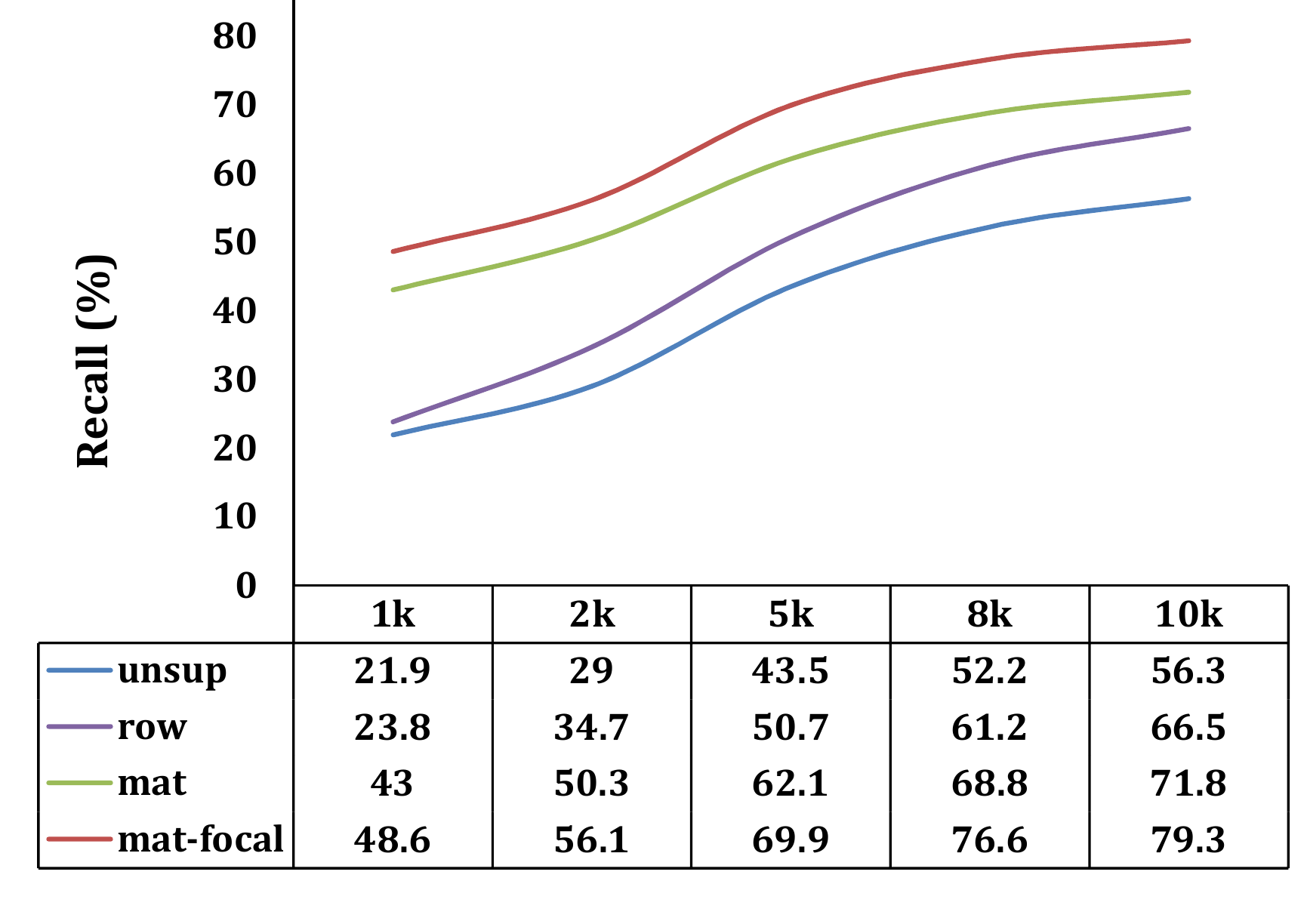}\\
\end{tabular}
\caption{Evaluating different target mass definitions. Results reported are relationship recall metric with varying top K, using the VOC07 test set. The supervision target $\mathcal{T}$ is constructed following Section 4.1 of main draft.}
\label{tab:mass_def_ablation}
\end{table}

\section{Target Mass Definition}
The definition of target mass, as in Section 3.2 of the main article, could have slightly different variations. We defined it as a summation over selected entries, in a matrix-scale. However, it is entirely possible to define such a summation over a row of matrix $\mathcal{W}$, when the softmax activation applied  is a row-wise operation. That is we only consider a row of matrix $\mathcal{W}$ during the softmax:
\begin{equation}
    \Tilde{\omega}_{ij}  = \frac{ \exp{\omega_{ij}} }{\sum_{x} \exp{\omega_{ix}} }
\end{equation}

We would build the target as we originally proposed, but compute the target mass in a row-wise manner and apply the affinity mass loss over the row-aggregated mass. For a given row $i$, we define its target mass as $\mathcal{M}_i$, and thus the affinity mass loss $\mathcal{L}_G$ can be written as 
\begin{equation}
    \mathcal{L}_G = - \sum_{i} (1 - \mathcal{M}_i)^{\gamma} \log{\mathcal{M}_i}.
\end{equation}

To justify the selected matrix-wise formulation that we proposed in the main draft, we provide the following ablation study, on relationship recovery recall metric using the VOC07 dataset. In the reported results here, we define the aforementioned row-wise target mass formulation as ``row'' and the matrix-wise version used in the main draft as ``mat'', but supervised with only the log-loss of $\mathcal{L}_G = - \log{\mathcal{M}}$. Lastly, to demonstrate the benefit of focal terms in the final loss form, which is
\begin{equation}
    \mathcal{L}_G = - (1 - \mathcal{M})^{\gamma} \log{\mathcal{M}},
\end{equation}
we define the matrix-wise supervision with focal term as ``mat-focal''. The recall measurement results are summarized in Table (\ref{tab:mass_def_ablation}). We emphasize that recall$@k$ reported here is always based on ranking the affinity weights post the matrix-wise softmax, ensuring fairness of the comparisons.

The results suggest that the affinity weights, when supervised using the affinity mass loss regardless of its form, are better than the unsupervised case (similar to Relation Networks \cite{hu2017relation}). Between different variations of target mass and loss forms, the choice of matrix-wise formulation with focal term gives the best results.

One can further simplify the definition of target mass to a single entry in matrix $\mathcal{W}$, and use a binary cross entropy loss over the Sigmoid activation of $\omega_{ij}$, which is $p_{ij} = \frac{1}{1 + \exp{ -\omega_{ij}} }$. The loss can be written as
\begin{equation}
    \mathcal{L}_G = -  \sum_{ij} \left[ \mathcal{T}_{ij} \log{p_{ij}} + 
    (1 - \mathcal{T}_{ij}) \log{ (1 - p_{ij}) } \right] ,
\end{equation}
where $\mathcal{T}_{ij}$ simply stands for the $i,j$-th entry of target matrix $\mathcal{T}$. Within multiple trials of a wide range of choices for the $\lambda$ defined in Section 3.4 of the main article, we found that this single entry based formulation does not converge to a sufficiently large target mass value and the recall metric is very close to the baseline unsupervised case, thus these results are inferior to the earlier formulations. 

In our loss design, the distinction between matrix softmax in affinity loss and row softmax in feature aggregation is essential, see Figure \ref{fig:structure_attn_module}. In affinity learning we care about accurately representing the strength of node-to-node connection. For instance, if a node has weak connection to all its neighbors, its edges should have relatively small weights. Following related work \cite{jiang2019semi, velivckovic2017graph, vaswani2017attention}, a row-wise softmax is applied in feature aggregation. This ensures a unified scaling of the aggregation result, so that a node with low affinity weights is not suppressed, and one with high weights is not dominant.

\end{document}